\documentclass[10pt,twocolumn,letterpaper]{article}

\usepackage[pagenumbers]{cvpr} 

\usepackage{times}
\usepackage{epsfig}
\usepackage{graphicx,scalerel}
\usepackage{amsmath}
\usepackage{amssymb}
\usepackage{booktabs}
\newcommand{\hide}[1]{}

\newcommand\wh[1]{\hstretch{2}{\hat{\hstretch{.5}{#1\mkern1mu}}}\mkern-1mu}
\usepackage{color}
\usepackage{mathtools}
\usepackage{multirow}
\usepackage{tipa}

\usepackage{bm} 
\definecolor{purple}{rgb}{0.65,0,0.65}
\definecolor{dark_green}{rgb}{0, 0.5, 0}
\definecolor{blueish}{rgb}{0.0, 0.3, .6}

\newcommand{\method}{SEARLE\xspace}

\newcommand\blfootnote[1]{%
  \begingroup
  \renewcommand\thefootnote{}\footnote{#1}%
  \addtocounter{footnote}{-1}%
  \endgroup
}

\usepackage[pagebackref,breaklinks,colorlinks]{hyperref}

\DeclareRobustCommand{\vect}[1]{\bm{#1}}
\usepackage[capitalize]{cleveref}
\crefname{section}{Sec.}{Secs.}
\Crefname{section}{Section}{Sections}
\Crefname{table}{Table}{Tables}
\crefname{table}{Tab.}{Tabs.}

\begin{document}

\title{Zero-Shot Composed Image Retrieval with Textual Inversion}

\author{\textsuperscript{$*$}Alberto Baldrati\textsuperscript{1,2} \and \textsuperscript{$*$}Lorenzo Agnolucci\textsuperscript{1} \and  Marco Bertini\textsuperscript{1} \and Alberto Del Bimbo\textsuperscript{1} \and
\textsuperscript{1} University of Florence - Media Integration and Communication Center (MICC) \\  \textsuperscript{2} University of Pisa \\ 
Florence, Italy - Pisa, Italy\\
{\tt\small [name.surname]@unifi.it}
}

\maketitle

\begin{abstract}
Composed Image Retrieval (CIR) aims to retrieve a target image based on a query composed of a reference image and a relative caption that describes the difference between the two images. The high effort and cost required for labeling datasets for CIR hamper the widespread usage of existing methods, as they rely on supervised learning. In this work, we propose a new task, Zero-Shot CIR (ZS-CIR), that aims to address CIR without requiring a labeled training dataset. Our approach, named zero-Shot composEd imAge Retrieval with textuaL invErsion (\method), maps the visual features of the reference image into a pseudo-word token in CLIP token embedding space and integrates it with the relative caption. To support research on ZS-CIR, we introduce an open-domain benchmarking dataset named Composed Image Retrieval on Common Objects in context (CIRCO), which is the first dataset for CIR containing multiple ground truths for each query. The experiments show that \method exhibits better performance than the baselines on the two main datasets for CIR tasks, FashionIQ and CIRR, and on the proposed CIRCO. The dataset, the code and the model are publicly available at \small{\href{https://github.com/miccunifi/SEARLE}{\url{https://github.com/miccunifi/SEARLE}}}. \blfootnote{\textsuperscript{$*$}~Equal contribution. Author ordering was determined by coin flip.} 
\end{abstract}

\section{Introduction}

\begin{figure}
    \centering
    \includegraphics[width=\columnwidth]{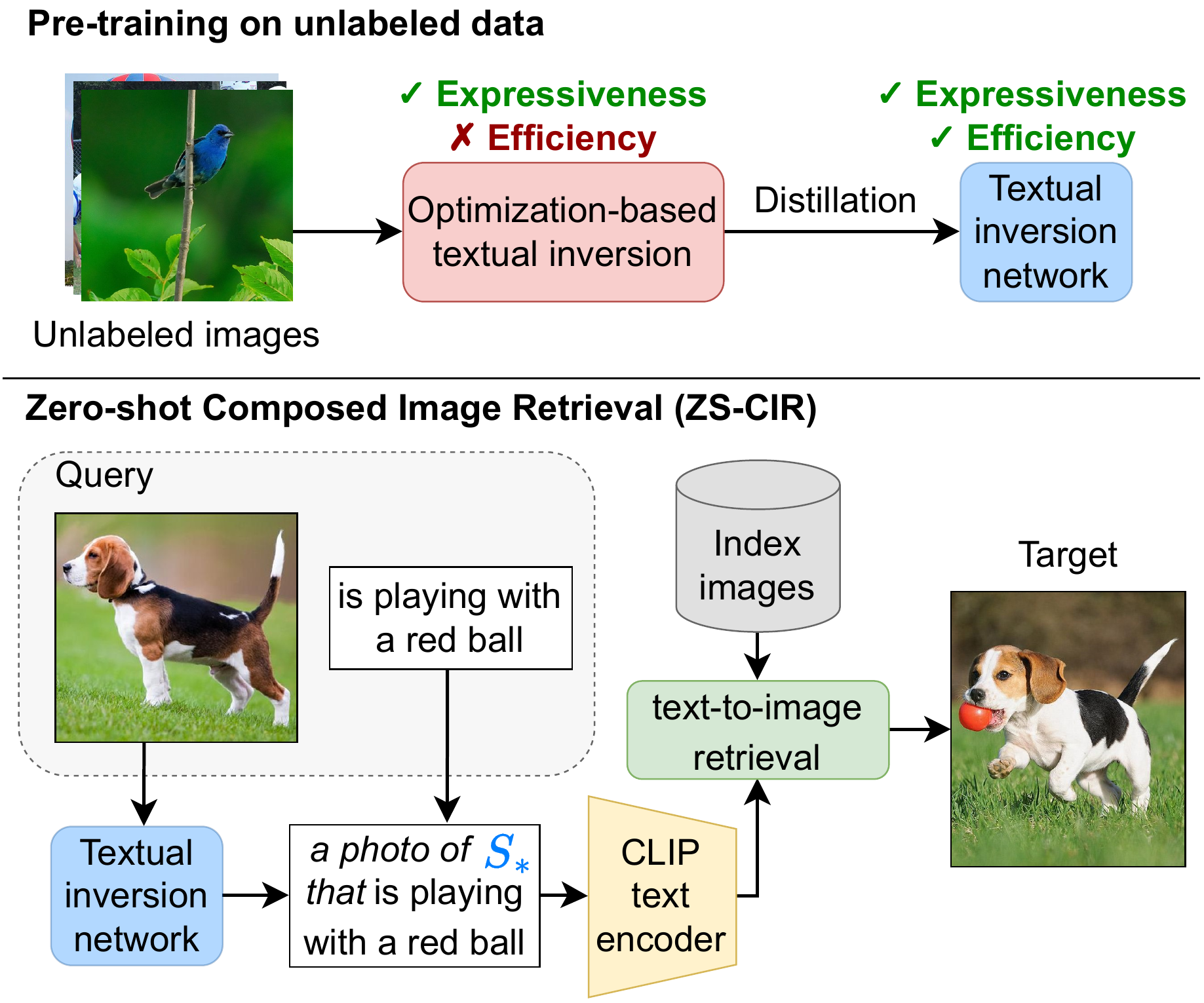}
    \caption{Workflow of our method. \textit{Top}: in the pre-training phase, we generate pseudo-word tokens of unlabeled images with an optimization-based textual inversion and then distill their knowledge to a textual inversion network. \textit{Bottom}: at inference time on ZS-CIR, we map the reference image to a pseudo-word $S_*$ and concatenate it with the relative caption. Then, we use CLIP text encoder to perform text-to-image retrieval.}
    \label{fig:intro}
\end{figure}

 Given a query composed of a reference image and a relative caption, Composed Image Retrieval (CIR)~\cite{vo2019composing, liu2021image} aims to retrieve target images that are visually similar to the reference one but incorporate the changes specified in the relative caption. The bi-modality of the query provides users with more precise control over the characteristics of the desired image, as some features are more easily described with language, while others can be better expressed visually. \Cref{fig:circo_example} shows some query examples.

CIR datasets consist of triplets $(I_r, T_r, I_t)$ composed of a reference image, a relative caption, and a target image, respectively. Creating a dataset for CIR is expensive as this type of data is not easily available on the internet, and generating it in an automated way is still very challenging. Thus, researchers must resort to manual labeling efforts. The manual process involves identifying pairs of reference and target images and writing a descriptive caption that captures the differences between them. This is a time-consuming and resource-intensive task, especially when creating large training sets. Current works tackling CIR~\cite{baldrati2022conditioned, baldrati2022effective, delmasartemis, liu2021image, lee2021cosmo} rely on supervision to learn how to combine the reference image and the relative caption. For instance, \cite{baldrati2022conditioned} proposes a fully-supervised two-stage approach that involves fine-tuning CLIP text encoder and training a combiner network. While current approaches for CIR have shown promising results, their reliance on expensive manually-annotated datasets limits their scalability and broader use in domains different from that of the datasets used for their training.

To remove the necessity of expensive labeled training data we introduce a new task, Zero-Shot Composed Image Retrieval (ZS-CIR). In ZS-CIR, the aim is to design an approach that manages to combine the reference image and the relative caption without the need for supervised learning. 

To tackle ZS-CIR, we propose an approach named zero-Shot composEd imAge Retrieval with textuaL invErsion (\method) \footnote{John Searle is an American philosopher who has studied the philosophy of language and how words are used to refer to specific objects.} that exploits the frozen pre-trained CLIP~\cite{radford2021learning} vision-language model. Our method reduces CIR to standard text-to-image retrieval by mapping the reference image into a learned pseudo-word which is then concatenated with the relative caption. The pseudo-word corresponds to a pseudo-word token residing in CLIP token embedding space. We refer to this mapping process with \textit{textual inversion}, following the terminology introduced in \cite{gal2022image}. \method involves pre-training a textual inversion network $\phi$ on an unlabeled image-only dataset. The training comprises two stages: an Optimization-based Textual Inversion (OTI) with a GPT-powered regularization loss to generate a set of pseudo-word tokens, and the distillation of their knowledge to $\phi$. After the training, we obtain a network $\phi$ that is able to perform textual inversion with a single forward pass. At inference time, given a query $(I_r, T_r)$, we use $\phi$ to predict the pseudo-word associated with $I_r$ and concatenate it to $T_r$. Then, we leverage the CLIP common embedding space to carry out text-to-image retrieval. \Cref{fig:intro} illustrates the workflow of the proposed approach.

Most available datasets for Composed Image Retrieval (CIR) focus on specific domains such as fashion~\cite{berg2010automatic, han2017automatic, wu2021fashion, guo2018dialog}, birds~\cite{forbes2019neural}, or synthetic objects \cite{vo2019composing}. To the best of our knowledge, the CIRR dataset \cite{liu2021image} is the only one that considers natural images in an open domain. However, CIRR suffers from two main issues. First, the dataset contains several false negatives, which could lead to an inaccurate performance evaluation. Second, the queries often do not consider the visual content of the reference image, making the task addressable with standard text-to-image techniques. Furthermore, existing CIR datasets have only one annotated ground truth image for each query. To address these issues and support research on ZS-CIR, we introduce an open-domain benchmarking dataset named Composed Image Retrieval on Common Objects in context (CIRCO)\footnote{CIRCO is pronounced as /\textipa{\textteshlig  \ignorespaces  irko}/.}, consisting of validation and test sets based on images from COCO \cite{lin2014microsoft}. Being a benchmarking dataset for ZS-CIR, the need for a large training set is removed, resulting in a significant reduction in labeling effort. To overcome the single ground truth limitation of existing CIR datasets, we propose a novel strategy that leverages \method to ease the annotation process of multiple ground truths. As a result, CIRCO is the first CIR dataset with multiple annotated ground truths, enabling a more comprehensive evaluation of CIR models. We release only the validation set ground truths of CIRCO and host an evaluation server to allow researchers to obtain performance metrics on the test set\footnote{\href{https://circo.micc.unifi.it/}{\url{https://circo.micc.unifi.it/}}}.

The experiments show that our approach obtains substantial improvements (up to 7\%) compared to the baselines on three different datasets: FashionIQ \cite{wu2021fashion}, CIRR \cite{liu2021image} and the proposed CIRCO.

Recently, a concurrent work \cite{saito2023pic2word} has independently proposed the same task as ours. In \cref{sec:related_work_textual_inversion} and \cref{sec:proposed_approach} we provide a detailed comparison illustrating the numerous differences from our approach, while in \cref{sec:experimental_results} we show that our method outperforms this work on all the test datasets.

Our contributions can be summarized as follows:
\vspace{-3pt}
\begin{itemize}
    \item We propose a new task, Zero-Shot Composed Image Retrieval (ZS-CIR), to remove the need for high-effort labeled data for CIR;
    \vspace{-3pt}
    \item We propose a novel approach, named \method, which employs a textual inversion network to tackle ZS-CIR by mapping images into pseudo-words. It involves two stages: an optimization-based textual inversion using a GPT-powered regularization loss and the training of the textual inversion network with a distillation loss;
    \vspace{-3pt} 
    \item We introduce CIRCO, an open-domain benchmarking dataset for ZS-CIR with multiple annotated ground truths and reduced false negatives. To ease the annotation process we propose to leverage \method;
    \vspace{-3pt}
    \item \method obtains significant improvements over baselines and competing methods achieving SotA on three different datasets: FashionIQ, CIRR, and the proposed CIRCO.
\end{itemize}


\section{Related Work}

\paragraph{Composed Image Retrieval} CIR belongs to the broader field of compositional learning, which has been extensively studied in various Vision and Language (V\&L) tasks, such as visual question answering~\cite{antol2015vqa}, image captioning~\cite{cornia2020meshed, hu2022scaling}, and image synthesis~\cite{rombach2022high, sahariaphotorealistic}. The goal of compositional learning is to generate joint-embedding features that capture relevant information from both text and visual domains.

The CIR task has been studied in various domains, such as fashion~\cite{berg2010automatic, han2017automatic, wu2021fashion, guo2018dialog}, natural images~\cite{forbes2019neural, liu2021image}, and synthetic images~\cite{vo2019composing}. It was first introduced in \cite{vo2019composing}, where the authors propose to compose the image-text features using a residual-gating method that aims to integrate the multimodal information. 
\cite{shin2021rtic} proposes a training technique that integrates graph convolutional networks with existing composition methods. \cite{lee2021cosmo} presents two different neural network modules that consider image style and content separately.
Recently, CLIP has been used to address CIR. \cite{baldrati2022effective} combines out-of-the-shelf image-text CLIP features using a combiner network, demonstrating their effectiveness. Later in \cite{baldrati2022conditioned}, the authors add a task-oriented fine-tuning step of the CLIP text encoder, achieving state-of-the-art performance. All of these approaches are supervised and require training on a CIR dataset to effectively learn to combine the multimodal information. In contrast, our method does not involve supervision and uses an unlabeled dataset for training.

\paragraph{Textual Inversion}\label{sec:related_work_textual_inversion}
In the field of text-to-image synthesis, mapping a group of images into a single pseudo-word has been proposed as a promising technique for generating highly personalized images~\cite{gal2022image, ruiz2023dreambooth, kumari2022multi, daras2022multiresolution}.
\cite{gal2022image} presents an approach for performing textual inversion using the reconstruction loss of a latent diffusion model~\cite{rombach2022high}.
In addition to textual inversion \cite{ruiz2023dreambooth} also fine-tunes a pre-trained text-to-image diffusion model. 

Besides personalized text-to-image synthesis, textual inversion has also been applied to image retrieval tasks~\cite{cohen2022this,saito2023pic2word, korbar2022personalised}. PALAVRA~\cite{cohen2022this} tackles personalized image retrieval, in which each query is composed of multiple images depicting a shared specific subject, and the goal is to retrieve images of such subject based on an input text in natural language. PALAVRA comprises two stages: the pre-training of a mapping function and a subsequent optimization.  It requires a labeled image-caption dataset for the pre-training and an input word concept for the optimization. On the contrary, we pre-train our textual inversion network on an unlabeled dataset and at inference time we do not need any additional inputs besides the reference image.
 
The most similar to our work is the concurrent Pic2Word \cite{saito2023pic2word}, which tackles ZS-CIR. Pic2Word relies on a textual inversion network trained on the 3M images of CC3M \cite{sharma2018conceptual} using only a cycle contrastive loss. Differently from this approach, we train our textual inversion network using only 3\% of the data and employing a weighted sum of distillation and regularization losses. The distillation loss exploits the information provided by a set of pre-generated tokens obtained through an optimization-based textual inversion. 

\paragraph{Knowledge Distillation}
Knowledge distillation is a machine learning technique where a simple model (student) is trained to mimic the behavior of a more complex one (teacher) by learning from its predictions \cite{hinton2015distilling}. This approach has been successfully applied to several computer vision tasks such as image classification \cite{hinton2015distilling, romero2014fitnets} and object detection \cite{chen2017learning}, achieving significant improvements in terms of model compression, speed, and performance. In our work, we refer to knowledge distillation as the process of transferring the knowledge acquired by a computationally expensive optimization method (teacher) to a light neural network (student). Specifically, we train a textual inversion network to mimic the output of an optimization-based textual inversion via a distillation loss. From another perspective, our light network can be interpreted as a surrogate model of the more resource-intensive optimization method.

\section{Proposed Approach} \label{sec:proposed_approach}

\begin{figure*}[!htb]
    \centering
    \includegraphics[width=\linewidth]{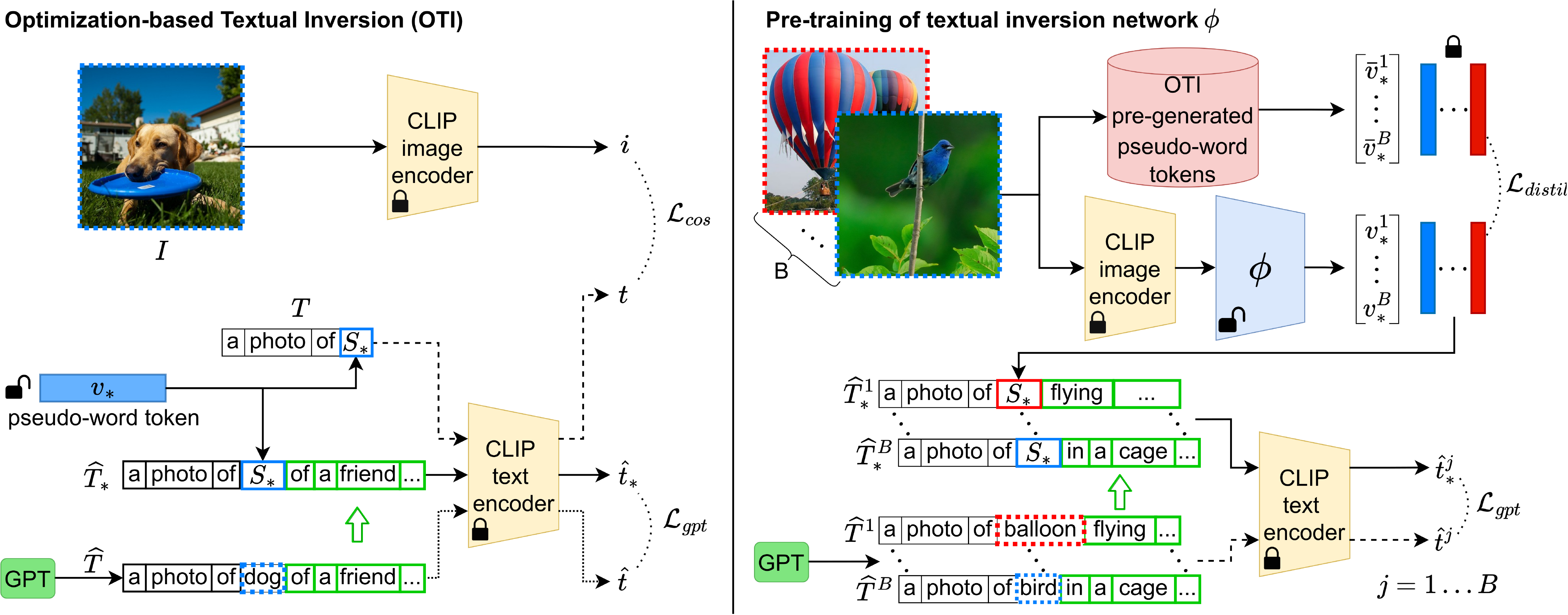}
    \vspace{-3.5ex}
    \caption{Overview of our approach. \textit{Left}: we generate a pseudo-word token $v_*$ from an image $I$ with an iterative optimization-based textual inversion. We force $v_*$ to represent the content of the image with a cosine loss $\mathcal{L}_{cos}$.
    We assign a concept word to $I$ with a CLIP zero-shot classification and feed the prompt ``a photo of \{concept\}" to GPT to continue the phrase, resulting in $\widehat{T}$. Let $S_*$ be the pseudo-word associated with $v_*$, we build $\widehat{T}_*$ by replacing in $\widehat{T}$ the concept with $S_*$. $\widehat{T}$ and $\widehat{T}_*$ are then employed for a contextualized regularization with $\mathcal{L}_{gpt}$. \textit{Right}: we train a textual inversion network $\phi$ on unlabeled images. Given a set of pseudo-word tokens pre-generated with OTI, we distill their knowledge to $\phi$ through a contrastive loss $\mathcal{L}_{distil}$. We regularize the output of $\phi$ with the same GPT-powered loss $\mathcal{L}_{gpt}$ employed in OTI. B represents the number of images in a batch.
    }
    \vspace{-4pt}
    \label{fig:teaser}
\end{figure*}

Our approach relies on CLIP (Contrastive Language-Image Pre-training (CLIP) \cite{radford2021learning}), a vision and language model that learns to align images and corresponding text captions in a common embedding space using a large-scale dataset. CLIP comprises an image encoder $\psi_{I}$ and a text encoder $\psi_{T}$. Given an image $I$, the image encoder extracts a feature representation $i = \psi_{I}(I) \in \mathbb{R}^{d}$, where $d$ is the size of CLIP embedding space. For a given text caption $T$, each tokenized word is mapped to the token embedding space $\mathcal{W}$ using a word embedding layer $E_w$. The text encoder $\psi_{T}$ is then applied to the token embeddings to produce the textual feature representation $t=\psi_{T}(E_w(T)) \in \mathbb{R}^{d}$. CLIP is trained to ensure that the same concepts expressed in an image or through text have similar feature representations.
 
Given a frozen pre-trained CLIP model, our approach, named \method, aims to generate a representation of the reference image that can be used as input to the CLIP text encoder. We achieve this goal by mapping the visual features of the image into a new token embedding belonging to $\mathcal{W}$. We refer to this token embedding as \textit{pseudo-word token}, since it is not associated with an actual word, but is rather a representation of the image features in the token embedding space. Our goal is twofold. First, we need to ensure that the pseudo-word token can accurately represent the content of the reference image. In other words, the text features of a basic prompt containing the pseudo-word need to be similar to the corresponding image features. Second, we need to make sure that such a pseudo-word can effectively integrate and communicate with the text of the relative caption.
Theoretically, a single image could be mapped to multiple pseudo-word tokens. In this work, we use a single one since \cite{gal2022image} shows that it is sufficient to encode the information of an image.

The first step of \method involves the pre-training of a textual inversion network $\phi$ on an unlabeled image-only dataset. The training is accomplished in two stages. First, we employ an Optimization-based Textual Inversion (OTI) method to iteratively generate a set of pseudo-word tokens leveraging a GPT-powered regularization loss. Second, we train $\phi$ by distilling the knowledge embedded in the pre-generated pseudo-word tokens. The $\phi$ network takes as input the features of an image extracted with CLIP image encoder and outputs the corresponding pseudo-word token in
a single forward pass. 

At inference time, CIR involves a query $(I_r, T_r)$ representing the input reference image and relative caption, respectively. We predict the pseudo-word token $v_*$ corresponding to the reference image as $ v_* = \phi(I_r)$. Let $S_*$ be the pseudo-word associated with $v_*$. To effectively integrate the visual information of $I_r$ with $T_r$, we construct the template ``a photo of $S_*$ that \textit{\{relative caption\}}" and extract its features using the CLIP text encoder. Notably, these text features comprise both textual and visual information, thus offering a multimodal representation of the reference image and its corresponding relative caption. Using the extracted text features, we perform a standard text-to-image retrieval by querying an image database. An overview of the workflow of our approach is illustrated in \cref{fig:intro}.

Conceptually, OTI and $\phi$ perform the same operation, \ie a textual inversion that maps the visual features of an image into a pseudo-word token. Therefore, we could directly employ OTI at inference time without the need for $\phi$. However, OTI requires a non-negligible amount of time to be carried out, while $\phi$ is significantly more efficient. Since OTI is proven to be powerful in generating effective pseudo-word tokens (see \cref{sec:experimental_results}), we propose to distill their knowledge into a feed-forward network. Our goal is to retain the expressive power of OTI while achieving a negligible inference time. In the following, we refer to \method when we employ $\phi$ to generate the pseudo-word token, and to \method-OTI when we directly use OTI for inference.  

\subsection{Optimization-based Textual Inversion (OTI)}\label{sec:optimization}
Given an image $I$, we adopt an optimization-based approach that performs the textual inversion by optimizing the pseudo-word token $v_* \in \mathcal{W}$ for a fixed amount of iterations. The left section of \cref{fig:teaser} shows an overview of OTI.

We start by randomly initializing the pseudo-word token $v_*$ and associating the pseudo-word $S_*$ to it. We build a template sentence $T$  such as ``a photo of $S_*$" and feed it to the CLIP text encoder $\psi_T$, obtaining $t = \psi_{T}(T)$. Similarly to \cite{cohen2022this}, we randomly sample $T$ from a pre-defined set of templates. Given an image $I$, we extract its features using the CLIP image encoder $\psi_{I}$, resulting in $i = \psi_{I}(I)$.

Since our goal is to obtain a pseudo-word token $v_*$ that encapsulates the informative content of $I$, we rely on CLIP common embedding space and minimize the gap between the image and text features. To achieve our aim, we leverage a cosine CLIP-based loss:
\begin{equation}
    \mathcal{L}_{cos} = 1 - \cos{(i, t)}
\end{equation}
 
However, $\mathcal{L}_{cos}$ alone is insufficient to generate a pseudo-word that can interact with other words in CLIP dictionary. Indeed, similarly to \cite{cohen2022this}, we observe that $\mathcal{L}_{cos}$ forces the pseudo-word token into sparse regions of CLIP token embedding space that are different from those observed during CLIP's training. An analogous effect has also been studied in GAN inversion works \cite{tov2021designing, zhu2020domain}. Consequently, the pseudo-word token is unable to effectively communicate with other tokens. To overcome this issue, we propose a novel regularization technique that constrains the pseudo-word token to reside on the CLIP token embedding manifold enhancing its reasoning capabilities (see \cref{sec:ablation_oti} for more details). First, we perform a zero-shot classification of the image $I$ relying on CLIP zero-shot capabilities. The vocabulary used to classify the images is taken from the $\sim$20K class names of the Open Images V7 dataset \cite{kuznetsova2020open}. In particular, we assign the most similar $k$ different class names to each image, where $k$ is a hyperparameter. We will refer to the class names used to classify the images as \textit{concepts}, \ie we associate each image to $k$ different concepts. Thanks to the zero-shot classification, different from \cite{cohen2022this}, we do not require the concepts as input.

Once we have a pool of concepts associated with the image, we generate a phrase using a lightweight GPT \cite{brown2020language} model. In each iteration of the optimization, we randomly sample one of the $k$ concepts associated with the image $I$ and feed the prompt ``a photo of \{concept\}" to GPT. Being an autoregressive generative model, GPT is capable of continuing this prompt in a meaningful way. For instance, given the concept ``dog", the GPT-generated phrase could be $\wh{T}$ = ``a photo of dog that was taken by his owner, who is a friend of mine". In practice, since the vocabulary is known a priori, we pre-generate all the GPT phrases for all the concepts in the vocabulary (see supplementary material for more details).
Starting from $\wh{T}$, we define $\wh{T}_*$ by simply replacing the concept with the pseudo-word $S_*$, obtaining $\wh{T}_*$ = ``a photo of $S_*$ that was taken\ldots".
We use the CLIP text encoder to extract the features of both phrases, ending up with  $\hat{t} = \psi_T(\wh{T})$ and $\hat{t}_* = \psi_T(\wh{T}_*)$.
Finally, we employ a cosine loss to minimize the gap between the features:
\begin{equation}
    \mathcal{L}_{gpt} = 1 -\cos{(\hat{t}, \hat{t}_*)}
\end{equation}
The idea behind this loss is to apply a contextualized regularization that pushes $v_*$ toward the concept while taking into account a broader context. Indeed, the GPT-generated phrases are more elaborated and thus similar to the relative captions used in CIR, compared to a generic pre-defined prompt. This way we enhance the ability of $v_*$ to interact with human-generated text such as the relative captions.

The final loss that we use for OTI is:
\begin{equation}\label{eq:loss_oti}
     \mathcal{L}_{OTI} =  \lambda_{cos} \mathcal{L}_{cos} + \lambda_{OTIgpt} \mathcal{L}_{gpt}
\end{equation}
where $\lambda_{cos}$ and $\lambda_{OTIgpt}$ are the loss weights.

\subsection{Textual Inversion Network $\vect{\phi}$ Pre-training}\label{sec:distillation}
We find OTI effective for obtaining pseudo-words that both encapsulate the visual information of the image and interact with actual words. However, being an iterative optimization-based method, it requires a non-negligible amount of time to be carried out. Therefore, we propose a method for training a textual inversion network $\phi$ capable of predicting the pseudo-word tokens in a single forward pass by distilling knowledge from a set of OTI pre-generated tokens. In other words, $\phi$ acts as a surrogate model of OTI, \ie a faster and less computationally heavy approximation of it. An overview of the pre-training phase is illustrated in the right part of \cref{fig:teaser}.

Our aim is to obtain a single model that can invert images of any domain and that does not need labeled data for training. In particular, we consider an MLP-based textual inversion network $\phi$ with three linear layers, each followed by a GELU \cite{hendrycks2016gaussian} activation function and a dropout layer. 

Given an unlabeled pre-training dataset $\mathcal{D}$, we start by applying OTI to each image. While this operation may be time-consuming, it is a one-time requirement, making it tolerable. We end up with a set of pseudo-word tokens $\overline{\mathcal{V}}_* = \{{\bar{v}_*^j}\}_{j=1}^N$, where $N$ is the number of images of $\mathcal{D}$. Our aim is to distill to $\phi$ the knowledge acquired by OTI and embedded in $\overline{\mathcal{V}}_*$. Starting from an image $I \in \mathcal{D}$, we extract its features using the CLIP visual encoder obtaining $i = \psi_{I}(I)$. We employ $\phi$ to predict the pseudo-word token $v_* = \phi(i)$.  We minimize the distance between the predicted pseudo-word token $v_*$ and its corresponding pre-generated token $\bar{v}_* \in \overline{\mathcal{V}}_*$ and, at the same time, maximize the discriminability of each token.
To this end, we employ a symmetric contrastive loss inspired by SimCLR \cite{chen2020simple, cohen2022this} as follows:
\begin{align} \label{eq:loss_distil}
    \mathcal{L}_{distil} = &\frac{1}{B} \sum^{B}_{k=1} -\log{\frac{e^{( c(\bar{v}_*^k, v_*^k) / \tau )}}{\sum\limits^B_{j=1} {e^{( c(\bar{v}_*^k, v_*^j) / \tau )}}+ {\sum\limits_{j\neq k} {e^{( c(v_*^k, v_*^j) / \tau )}}} }}\nonumber \\ 
    &- \log{\frac{e^{( c(v_*^k, \bar{v}_*^k) / \tau )}}{\sum\limits^B_{j=1} {e^{( c(v_*^k, \bar{v}_*^j) / \tau )}}+ {\sum\limits_{j\neq k} {e^{( c(\bar{v}_*^k, \bar{v}_*^j) / \tau )}}} }} 
\end{align}

Here, $c(\cdot)$ denotes the cosine similarity, $B$ is the number of images in a batch, and $\tau$ is a temperature hyperparameter.

To regularize the training of $\phi$ we follow the same technique described in \cref{sec:optimization} and rely on $\mathcal{L}_{gpt}$.

The final loss used to update the weights of $\phi$ is
\begin{equation}\label{eq:loss_phi}    
    \mathcal{L}_{\phi} = \lambda_{distil} \mathcal{L}_{distil} +  \lambda_{\phi gpt} \mathcal{L}_{gpt}
\end{equation}
where $\lambda_{distil}$ and $\lambda_{\phi gpt}$ are the loss weights.

The training of our textual inversion network $\phi$ is fully unsupervised as we do not rely on any labeled data. Indeed, we utilize raw images, different from \cite{cohen2022this}, which also needs captions. In particular, we employ the unlabeled test split of the ImageNet1K \cite{russakovsky2015imagenet} dataset as $\mathcal{D}$ to pre-train $\phi$. It contains 100K images without any given label. In comparison to PALAVRA \cite{cohen2022this} and Pic2Word \cite{saito2023pic2word}, our method utilizes significantly fewer data, approximately the 10\%, and the 3\%, respectively. We chose this dataset as it includes real-world images with a high variety of subjects. We believe that other similar datasets could serve our purpose.

\section{CIRCO dataset} \label{sec:circo_dataset}

\begin{figure}
    \centering
    \includegraphics[width=\columnwidth]{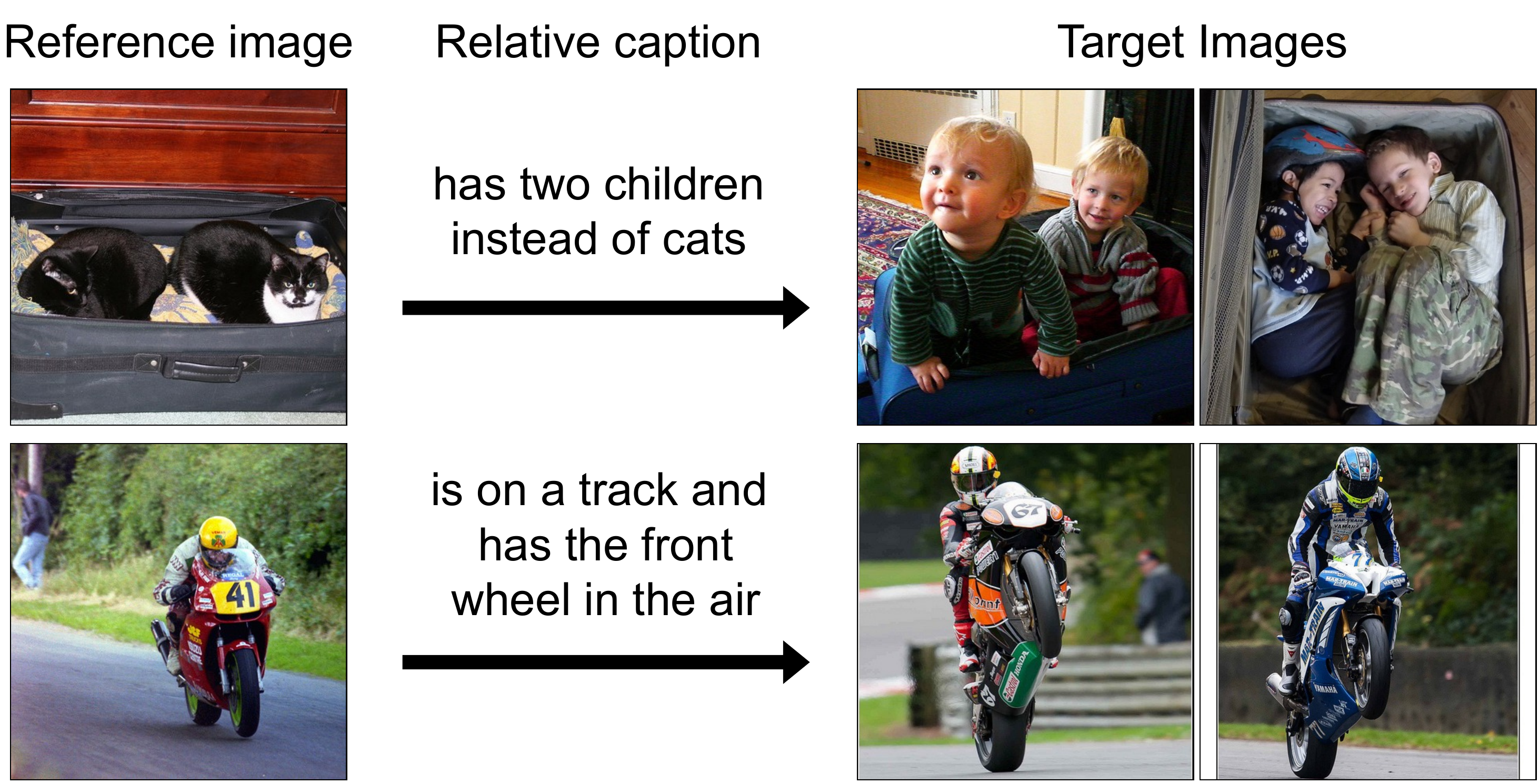}
    \vspace{-3.5ex}
    \caption{Examples of CIR queries and ground truths in CIRCO.}
    \vspace{-4pt}
    \label{fig:circo_example}
\end{figure}

We recall that CIR datasets consist of triplets $(I_r, T_r, I_t)$ composed of a reference image, relative caption, and target image (\ie the ground truth), respectively. 

Existing datasets contain several false negatives, \ie images that could be potential ground truths for the query but are not labeled as such. Indeed, since each query triplet contains only a target, all the other images are considered negatives. In addition, most datasets revolve around specialized domains such as fashion~\cite{berg2010automatic, han2017automatic, wu2021fashion, guo2018dialog}, birds~\cite{forbes2019neural}, or synthetic objects~\cite{vo2019composing}.
To the best of our knowledge, CIRR \cite{liu2021image} is the only dataset based on real-life images in an open domain. During the data collection process, CIRR builds sets of 6 visually similar images in an automated way. Then, the queries are created such that the reference and the target images belong to the same set and so as to avoid the presence of false negatives within the set. The flaw with this approach is that it does not guarantee the absence of false negatives in the whole dataset. Furthermore, despite the visual similarity, the difference between images belonging to the same set can possibly be not easily describable with a relative caption and require an absolute description. This reduces the importance of the visual information of the reference image and makes the retrieval addressable with standard text-to-image methods (see \cref{sec:cirr_results} for more details).

To address these issues, we introduce an open-domain benchmarking dataset named Composed Image Retrieval on Common Objects in context (CIRCO). It is based on real-world images from COCO 2017 unlabeled set \cite{lin2014microsoft} and is the first dataset for CIR with multiple ground truths. Contrary to CIRR, we start from a single pair of visually similar images and write a relative caption. In addition, we provide an auxiliary annotation with the shared characteristics of the reference and target images to clarify ambiguities. Then, we propose to employ our approach to retrieve the top 100 images according to the query and combine them with the top 50 images most visually similar to the target one. Finally, we select the images that are valid matches for the query. We estimate that this approach allows us to reduce the percentage of missing ground truths in the dataset to less than 10\%. CIRCO comprises a total of 1020 queries, randomly divided into 220 and 800 for the validation and test set, respectively, with an average of 4.53 ground truths per query. We use all the 120K images of COCO as the index set, thus providing significantly more distractors than the 2K images of CIRR test set. \Cref{fig:circo_example} shows some query examples. More details on the CIRCO dataset and a more comprehensive comparison with CIRR are provided in the supplementary material.

To mitigate the problem of false negatives, most works evaluate the performance using Recall@K, with K set to quite large values (\eg 10, 50 \cite{wu2021fashion}), thus making a fine-grained analysis of the models difficult. CIRR addresses the issue by employing also Recall$_{\text{Subset}}@K$, which considers only the images in the same set of the reference and target ones. Thanks to our multiple ground truths, we can rely on a more fine-grained metric such as mean Average Precision (mAP), which takes into account also the ranks in which the ground truths are retrieved. In particular, we use mAP$@K$, with K ranging from small to quite large values.


\section{Experimental Results} \label{sec:experimental_results}
We test our approach following the standard evaluation protocol \cite{baldrati2022conditioned, liu2021image} on three datasets: FashionIQ \cite{wu2021fashion}, CIRR \cite{liu2021image} and the proposed CIRCO. In particular, we employ the three categories of FashionIQ validation split and the test sets of CIRR and CIRCO. We introduce two variants of our approach: \method, based on CLIP ViT-B/32, and \method-XL, using CLIP ViT-L/14 as the backbone.
In the following, we refer to ViT-B/32 and ViT-L/14 as B/32 and L/14, respectively. For the sake of space, we provide the implementation details and the qualitative results in the supplementary material.

\subsection{Quantitative Results}

\begin{table*}[!ht]
\centering
  \resizebox{0.8\linewidth}{!}{ 
  \begin{tabular}{clcccccccc} 
  \toprule
  \multicolumn{1}{c}{} & \multicolumn{1}{c}{} & \multicolumn{2}{c}{Shirt} & \multicolumn{2}{c}{Dress} &\multicolumn{2}{c}{Toptee} &\multicolumn{2}{c}{Average} \\
  \cmidrule(lr){3-4}
  \cmidrule(lr){5-6}
  \cmidrule(lr){7-8}
  \cmidrule(lr){9-10}
  \multicolumn{1}{c}{Backbone} & \multicolumn{1}{l}{Method} & R$@10$ & R$@50$ & R$@10$ & R$@50$ & R$@10$ & R$@50$ & R$@10$ & R$@50$ \\ 
  \midrule
  \multirow{7}{*}{B/32} & Image-only & 6.92 & 14.23 & 4.46 & 12.19 & 6.32 & 13.77 & 5.90 & 13.37 \\
  & Text-only & 19.87 & 34.99 & 15.42 & 35.05 & 20.81 & 40.49 & 18.70 & 36.84 \\ 
  & Image + Text & 13.44 & 26.25 & 13.83 & 30.88 & 17.08 & 31.67 & 14.78 & 29.60 \\
  & Captioning & 17.47 & 30.96 & 9.02 & 23.65 & 15.45 & 31.26 & 13.98 & 28.62 \\
  & PALAVRA \cite{cohen2022this} & 21.49 & 37.05 & 17.25 & 35.94 & 20.55 & 38.76 & 19.76 & 37.25 \\
  & \textbf{\method-OTI} & \textbf{25.37} & \underline{41.32} & \underline{17.85} & \textbf{39.91} & \underline{24.12} & \underline{45.79} & \underline{22.44} & \underline{42.34} \\
  & \textbf{\method} & \underline{24.44} & \textbf{41.61} & \textbf{18.54} & \underline{39.51} & \textbf{25.70} & \textbf{46.46} & \textbf{22.89} & \textbf{42.53}  \\ 
  \midrule[.02em]
  \multirow{3}{*}{L/14} & Pic2Word$^{\dagger}$ \cite{saito2023pic2word} & 26.20 & 43.60 & 20.00 & 40.20 & 27.90 & 47.40 & 24.70 & 43.70 \\
  & \textbf{\method-XL-OTI} & \textbf{30.37} & \textbf{47.49} & \textbf{21.57} & \textbf{44.47} & \textbf{30.90} & \textbf{51.76} & \textbf{27.61} & \textbf{47.90} \\
  & \textbf{\method-XL} & \underline{26.89} & \underline{45.58} & \underline{20.48} & \underline{43.13} & \underline{29.32} & \underline{49.97} & \underline{25.56} & \underline{46.23} \\ 
  \bottomrule
  \end{tabular}}
  \caption{Quantitative results on FashionIQ validation set. Best and second-best scores are highlighted in bold and underlined, respectively. $^{\dagger}$ indicates results from the original paper.}
  \label{tab:fashioniq_val}
\end{table*}

\begin{table*}[!ht]
 \centering
  \resizebox{0.8\linewidth}{!}{ 
  \begin{tabular}{clccccccc} 
  \toprule
  \multicolumn{1}{c}{} & \multicolumn{1}{c}{} & \multicolumn{4}{c}{Recall$@K$} & \multicolumn{3}{c}{Recall$_{\text{Subset}}@K$} \\
  \cmidrule(lr){3-6}
  \cmidrule(lr){7-9}
  \multicolumn{1}{l}{Backbone} & \multicolumn{1}{l}{Method} & $K = 1$ & $K = 5$ & $K = 10$ & $K = 50$ & $K = 1$ & $K = 2$ & $K = 3$ \\ 
  \midrule
  \multirow{7}{*}{B/32} & Image-only & 6.89  & 22.99  & 33.68 & 59.23 & 21.04 & 41.04 & 60.31 \\ 
  & Text-only & 21.81 & 45.22 & 57.42 & 81.01 & \textbf{62.24} & \textbf{81.13} & \textbf{90.70} \\ 
  & Image + Text & 11.71 & 35.06 & 48.94 & 77.49 & 32.77  & 56.89 & 74.96 \\
  & Captioning & 12.46 & 35.04 & 47.71 & 77.35 & 42.94  & 65.49 & 80.36 \\
  & PALAVRA \cite{cohen2022this} & 16.62 & 43.49 & 58.51 & 83.95 & 41.61 & 65.30 & 80.94 \\
  & \textbf{\method-OTI} & \textbf{24.27} & \underline{53.25} & \underline{66.10} & \underline{88.84} & 54.10 & 75.81 & 87.33 \\
  & \textbf{\method} & \underline{24.00}  & \textbf{53.42} & \textbf{66.82} & \textbf{89.78} & \underline{54.89} & \underline{76.60} & \underline{88.19}  
  \\ \midrule[.02em]
  \multirow{3}{*}{L/14} & Pic2Word$^{\dagger}$ \cite{saito2023pic2word} & 23.90 & 51.70 & 65.30 & 87.80 & -- & -- & -- \\
  & \textbf{\method-XL-OTI} & \textbf{24.87} & \underline{52.31} & \textbf{66.29} & \underline{88.58} & \textbf{53.80} & \underline{74.31} & \underline{86.94}  \\
  & \textbf{\method-XL} & \underline{24.24} & \textbf{52.48} & \textbf{66.29} & \textbf{88.84} & \underline{53.76} & \textbf{75.01} & \textbf{88.19} 
  \\
  \bottomrule
  \end{tabular}}
  \caption{Quantitative results on CIRR test set. Best and second-best scores are
highlighted in bold and underlined, respectively. $^{\dagger}$ indicates results from the original paper. -- denotes results not reported in the original paper.}
  \label{tab:cirr_test}
\end{table*}

We compare our approach with several zero-shot baselines and competing methods, including: 1) \textit{Text-only}: the similarity is computed using only the CLIP features of the relative caption; 2) \textit{Image-only}: retrieves the most similar images to the reference one; 3) \textit{Image + Text}: the CLIP features of the reference image and the relative caption are summed together; \textit{4) Captioning}: we substitute the pseudo-word token with the caption of the reference image generated with a pre-trained captioning model~\cite{yu2022coca}\footnote{\href{https://huggingface.co/laion/CoCa-ViT-B-32-laion2B-s13B-b90k}{https://huggingface.co/laion/CoCa-ViT-B-32-laion2B-s13B-b90k}}
5) \textit{PALAVRA} \cite{cohen2022this}: a textual inversion-based two-stage approach with a pre-trained mapping function and a subsequent optimization of the pseudo-word token; 6) \textit{Pic2Word} \cite{saito2023pic2word}: forward-only method employing a pre-trained textual inversion network.

For PALAVRA, we use CLIP B/32 as the backbone following the original paper, while for Pic2Word, we report the results provided by the authors when available.
Considering our approach, we provide the results of both \method and \method-OTI.

\paragraph{FashionIQ}
\Cref{tab:fashioniq_val} shows the results on FashionIQ. With the B/32 backbone, \method achieves comparable performance with \method-OTI. It is worth noting that \method provides a significant efficiency gain without compromising performance. Our approach outperforms the baselines in both versions. In particular, the enhancement over Captioning underscores
that the pseudo-word token embeds more information than the actual words comprising the generated caption. Considering the L/14 backbone, \method-XL significantly improves over Pic2Word, up to a 7\% gain in the Recall@50 for the Dress category. We recall that the two approaches are directly comparable as they both rely on a single forward of a pre-trained network with no subsequent optimization, but our model is trained with 3\% of the data. However, we notice a gap with the performance obtained by \method-XL-OTI. We suppose it is due to the very narrow domain of FashionIQ, which is quite different from the natural images of the pre-training dataset we use for training $\phi$. To support our theory, we trained a version of $\phi$ using the FashionIQ training set as the pre-training dataset, obtaining an average R$@10$ and R$@50$ of 27.95 and 49.24 in the validation set, respectively. These results are comparable with \method-XL-OTI, confirming our hypothesis. More details are provided in the supplementary material.

\paragraph{CIRR} \label{sec:cirr_results}
In \cref{tab:cirr_test} we report the results for CIRR test set. The Text-only baseline obtains the best performance on Recall$_{\text{Subset}}$ and outperforms Image-only and Image+Text in the global metrics. These results highlight a major flaw in CIRR: the relative captions are often not actually relative in practice. Specifically, we find that the reference image may not provide useful information for retrieval, and may even have a detrimental effect, as also observed in \cite{saito2023pic2word}. This is especially true when considering only the subset of images that comprises the reference and target ones, as the visual information is very similar. Indeed, the Recall$_{\text{Subset}}$ results of Image-only correspond to random guessing as the retrieval in the subset involves only five images. 

In this dataset, we notice that for both backbones the results obtained by our approach with OTI and $\phi$ are comparable, showing the effectiveness of our distillation process. It is worth noticing that there is no performance gap between the B/32 and L/14 versions, and in some cases, the B/32 even outperforms the L/14. We improve over PALAVRA and Pic2Word when using the same backbones. Unfortunately, we can not compare our performance on Recall$_{\text{Subset}}$ with Pic2Word as the authors do not report their results.

\paragraph{CIRCO}
\Cref{tab:circo_test} shows the results on CIRCO test set. First, we can notice how, contrary to FashionIQ and CIRR, Image+Text achieves better results than Image-only and Text-only. This shows how CIRCO comprises queries in which the reference image and the relative caption are equally important to retrieve the target images. Second, \method significantly improves over all the baseline methods, even outperforming Pic2Word, which employs a larger backbone. \method-XL would achieve the best results, but we do not consider them completely fair as it was employed to retrieve the images that the multiple ground truths were selected from. Still, we report them for completeness and as a baseline for future works that will use our dataset for testing.

\begin{table}[!ht]
  \centering
  \Huge
  \resizebox{\linewidth}{!}{ 
  \begin{tabular}{clcccc} 
  \toprule
  \multicolumn{1}{c}{} & \multicolumn{1}{c}{} & \multicolumn{4}{c}{mAP$@K$} \\
  \cmidrule(lr){3-6}
  \multicolumn{1}{l}{Backbone} & \multicolumn{1}{l}{Method} & $K=5$ & $K=10$ & $K=25$ & $K=50$ \\ \midrule
  \multirow{7}{*}{B/32} & Image-only & 1.34 & 1.60 & 2.12 & 2.41 \\
  & Text-only & 2.56 & 2.67 & 2.98 & 3.18 \\
  & Image + Text & 2.65 & 3.25 & 4.14 & 4.54 \\
  & Captioning & 5.48 & 5.77 & 6.44 & 6.85 \\
  & PALAVRA \cite{cohen2022this} & 4.61 & 5.32 & 6.33 & 6.80 \\
  & \textbf{\method-OTI} & \underline{7.14} & \underline{7.83} & \underline{8.99} & \underline{9.60} \\
   & \textbf{\method} & \textbf{9.35} & \textbf{9.94} & \textbf{11.13} & \textbf{11.84} \\ \midrule[.02em]
  \multirow{3}{*}{L/14} & Pic2Word~\cite{saito2023pic2word} & 8.72 & 9.51& 10.64&11.29 \\
  & \textbf{\method-XL-OTI} & \underline{10.18} & \underline{11.03} & \underline{12.72} & \underline{13.67} \\
  & \textbf{\method-XL} & \textbf{11.68}  & \textbf{12.73} & \textbf{14.33} & \textbf{15.12} \\
  \bottomrule
  \end{tabular}}
  \caption{Quantitative results on CIRCO test set. Best and second-best scores are highlighted in bold and underlined, respectively.}
  \label{tab:circo_test}
\end{table}

\subsection{Ablation Studies}
We conduct ablation studies to evaluate the individual contributions of the components in our approach. To avoid confounding effects, we assess the two main components of the proposed method separately. Specifically, we evaluate the textual inversion network $\phi$ while keeping fixed the set of OTI pre-generated tokens obtained with the method described in \cref{sec:optimization}. As $\phi$ distills their knowledge, we assume that the more informative they are (\ie the better OTI performs), the better the results obtained by $\phi$ will be.

We perform the ablation studies on CIRR and FashionIQ validation sets and report the results for the main evaluation metrics. In particular, for FashionIQ we report the average scores. For simplicity, we consider only the version of our approach with the B/32 backbone.

\begin{table}[!t]
  \centering
  \Large
  \resizebox{\linewidth}{!}{ 
  \begin{tabular}{clccccc} 
  \toprule
  \multicolumn{1}{c}{} & \multicolumn{1}{c}{} & \multicolumn{3}{c}{CIRR} & \multicolumn{2}{c}{FashionIQ}\\
  \cmidrule(lr){3-5}
  \cmidrule{6-7}
  \multicolumn{1}{l}{Abl.} & \multicolumn{1}{l}{Method} & R$@1$ & R$@5$ & R$@10$ & R$@10$ & R$@50$ \\ \midrule
   \multirow{4}{*}{OTI} & w/o GPT reg & \underline{21.63} & \underline{50.51} & \underline{64.07} & \underline{21.34} & 39.70 \\
  & random reg & 21.09 & 50.42 & 63.84 & 20.90 & \underline{40.24} \\
  & w/o reg & 19.30 & 46.81 & 59.96 & 17.86 & 35.99 \\
  & \textbf{\method-OTI} & \textbf{23.54} & \textbf{53.93} & \textbf{67.69} & \textbf{22.44} & \textbf{42.34} \\ \midrule[.02em]

  \multirow{4}{*}{$\phi$}  & cos distil & \underline{23.75} & \underline{53.19} & \underline{67.21} & 21.59 & 39.41 \\
  & w/o distil & 22.24 & 50.06 & 62.71 & 19.41 & 38.39 \\
  & w/o reg & 22.41 & 53.00 & 66.90 & \underline{22.83} & \underline{42.02} \\
  & \textbf{\method} & \textbf{25.09}  & \textbf{55.18} & \textbf{68.79} & \textbf{22.89} & \textbf{42.53} \\
  \bottomrule
  \end{tabular}}
  \caption{Ablation studies on CIRR and FashionIQ validation sets. For FashionIQ, we consider the average recall. Best and second-best scores are highlighted in bold and underlined, respectively.} 
  \label{tab:ablation}
\end{table}

\paragraph{Optimization-based textual inversion (OTI)}\label{sec:ablation_oti}
We ablate the regularization loss used during the optimization process: 1) \textit{w/o GPT reg}: we regularize with a prompt comprising only the concept word, without the GPT-generated suffix; 2) \textit{random reg}: we additionally substitute the concept word with a random word; 3) \textit{w/o reg}: we completely remove the regularization loss.

The upper section of \cref{tab:ablation} shows the results. As a different loss corresponds to a different speed of convergence, for each ablation experiment we use a tailored number of optimization iterations and report the best performance. 
We notice that some kind of regularization is essential to make the pseudo-word tokens reside on the CLIP token embedding manifold and communicate effectively with CLIP vocabulary tokens. Specifically, the proposed GPT-based regularization loss allows the pseudo-word tokens to interact with text resembling human-written text, thus enhancing their communication with the relative caption and the performance of retrieval. This is especially true for CIRR, as the relative captions are more elaborated and have a more varied vocabulary.

\paragraph{Textual inversion network $\vect{\phi}$} \label{sec:ablation_phi}
We ablate the losses used during the pre-training: 1) \textit{cos distil}: we employ a cosine distillation loss instead of a contrastive one;  2) \textit{w/o distil}: we replace $\mathcal{L}_{distil}$ with the cycle contrastive loss employed by \cite{cohen2022this}, which directly considers the image and text features; 3) \textit{w/o reg}: we remove the $\mathcal{L}_{gpt}$ regularization loss.

We report the results in the lower part of \cref{tab:ablation}. The contrastive version of the distillation loss proves to be more effective than the cosine one. Compared to the cycle contrastive loss, our distillation-based loss achieves significantly superior performance, showing how learning from OTI pre-generated tokens is more fruitful than from raw images. Finally, although the pre-generated pseudo-word tokens are already regularized, we notice that our GPT-based regularization loss is still beneficial for training $\phi$.


\section{Conclusion}
In this paper we introduce a new task, Zero-Shot Composed Image Retrieval (ZS-CIR), that aims to address CIR without the need for an expensive labeled training dataset. Our approach, named \method, involves the pre-training of a textual inversion network that leverages a distillation loss to retain the expressive power of an optimization-based method while achieving a significant efficiency gain. We also present a new open-domain benchmarking dataset for CIR, Composed Image Retrieval on Common Objects in context (CIRCO). CIRCO is the first dataset for CIR that contains multiple ground truths for each query. Both versions of our approach achieve superior performance over baselines and competing methods on popular datasets such as CIRR and FashionIQ and on the proposed CIRCO.

In future work, we plan to investigate the potential of our method in the personalized image generation task. In particular, we believe that the proposed GPT-based regularization loss could improve the ability of a generative model to consider input text for synthesizing personalized objects.

\paragraph{Acknowledgments}
This work was partially supported by the European Commission under European Horizon 2020 Programme, grant number 101004545 - ReInHerit.

{\small
\bibliographystyle{ieee_fullname}
\bibliography{egbib}
}

\clearpage
\appendix
\section*{Appendix}

\section{Implementation Details}
For the Optimization-based Textual Inversion (OTI), we perform 350 iterations with a learning rate of $2e\!-\!2$. We set the loss weights $\lambda_{cos}$ and $\lambda_{OTIgpt}$ in \cref{eq:loss_oti} to 1 and 0.5, respectively. We adopt an exponential moving average with 0.99 decay. Regarding the training of the textual inversion network $\phi$, we train for 100 and 50 epochs for \method and \method-XL, respectively, with a learning rate of $1e\!-\!4$ and a batch size of 256. We set the loss weights $\lambda_{distil}$ and $\lambda_{\phi gpt}$ in \cref{eq:loss_phi} to 1 and 0.75, respectively. The temperature $\tau$ in \cref{eq:loss_distil} is set to 0.25. For both OTI and $\phi$, we employ the AdamW optimizer \cite{loshchilov2018decoupled} with weight decay 0.01. 
During OTI we set the number of concept words $k$ associated with each image to 15, while during the training of $\phi$ to 150. We tune each hyperparameter individually with a grid search on the CIRR validation set.
With a single A100 GPU, OTI for \method-XL takes $\sim$30 seconds for a single image and $\sim$1 second per image with batch size 256. The training of $\phi$ for \method-XL takes 6 hours in total on a single A100 GPU. Throughout all the experiments, we use the pre-processing technique proposed in \cite{baldrati2022effective}. We use mixed precision to save memory and increase computational efficiency. For retrieval, we normalize both the query and index set features to have a unit $L_2$-norm.

To generate the phrases used for the regularization with $\mathcal{L}_{gpt}$, we employ the GPT-Neo-2.7B model with 2.7 billion parameters developed by EleutherAI. For each of the 20,932 class names of the Open Images V7 dataset \cite{kuznetsova2020open}, we generate 256 phrases a priori with a temperature of 0.5, constraining the length to a maximum of 35 tokens. The whole process takes approximately 12 hours to complete on a single NVIDIA A100 GPU. We need to perform this operation only once, making the time requirements tolerable.

Since the FashionIQ dataset provides two relative captions for each triplet, during inference, we concatenate them using the conjunction ``and". To ensure our approach remains unaffected by the order of concatenation, we employ both potential concatenation orders and afterward average the resulting features.

\subsection{$\vect{\phi}$ Architecture}
\Cref{tab:phi_architecture} illustrates the details of the architecture of the textual inversion network $\phi$. For the B/32 backbone, the dimension of the CLIP feature space and token embedding space $\mathcal{W}$, respectively $d$ and $d_w$, are both equal to 512. For the L/14 backbone $d$ and $d_w$ both equal 768. 

\begin{table}
    \centering
    \begin{tabular}{cc}
        Layer & Module \\ \midrule
        Input & nn.Linear($d$, $d*4$) \\
        GELU & nn.GELU \\
        Dropout & nn.Dropout(0.5) \\
        Hidden & nn.Linear($d*4$, $d*4$) \\
        GELU & nn.GELU \\
        Dropout & nn.Dropout(0.5) \\
        Output & nn.Linear($d*4$, $d_w$) \\ \bottomrule
    \end{tabular}
    \caption{Pytorch-style description of the textual inversion network $\phi$. $d$ and $d_w$ represent the dimension of the CLIP feature space and token embedding space $\mathcal{W}$, respectively.}
    \vspace{-3ex}
    \label{tab:phi_architecture}
\end{table}

\begin{figure}
  \centering
  \begin{subfigure}{\linewidth}
    \frame{\includegraphics[width=\linewidth]{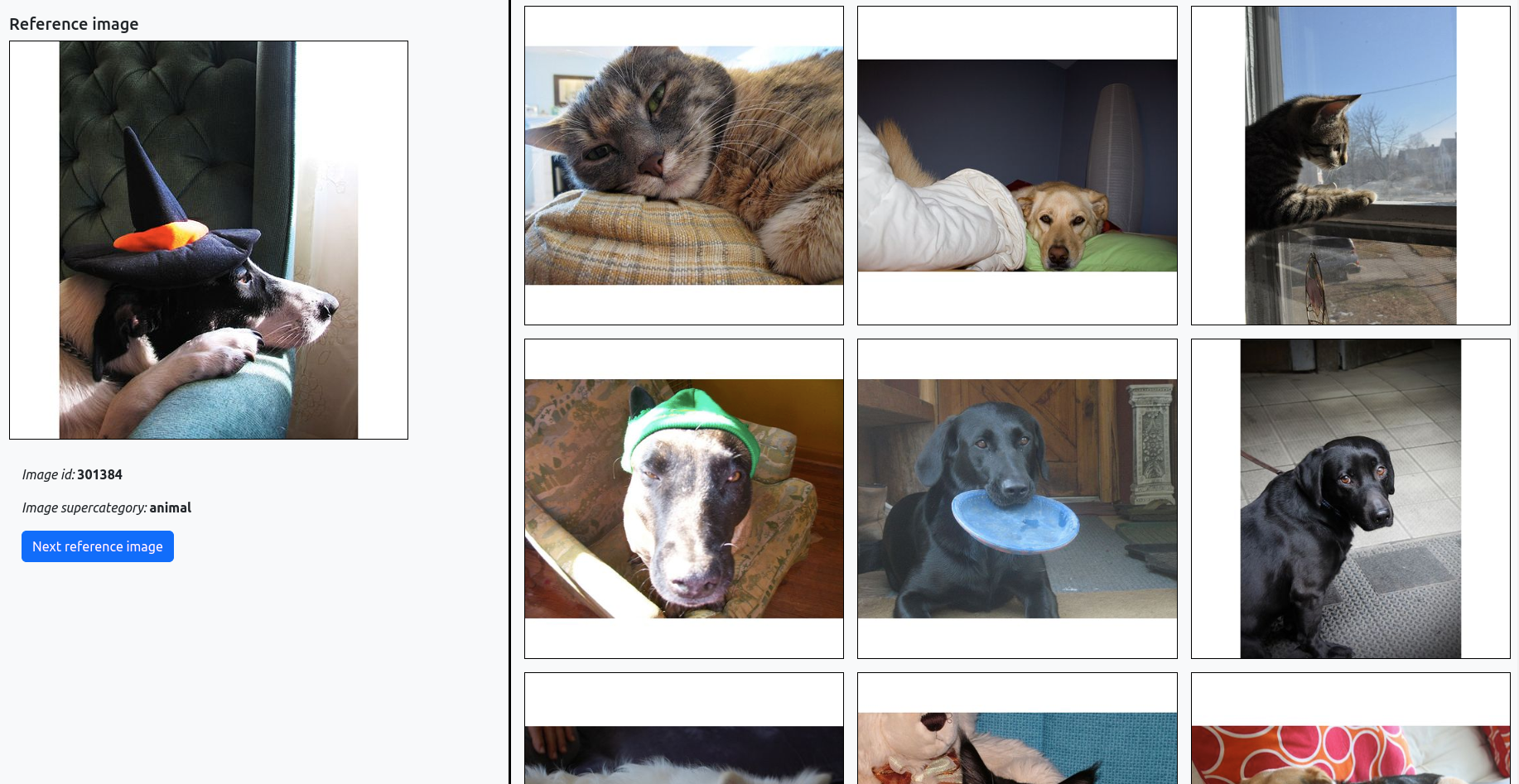}}
    \caption{Interface for selecting the reference-target image pair. On the left, there is a randomly sampled reference image and a button to skip it. On the right, a gallery of images displays the candidate target images. \vspace{10pt}}
    \label{fig:annotator_index}
  \end{subfigure}
  \begin{subfigure}{\linewidth}
    \frame{\includegraphics[width=\linewidth]{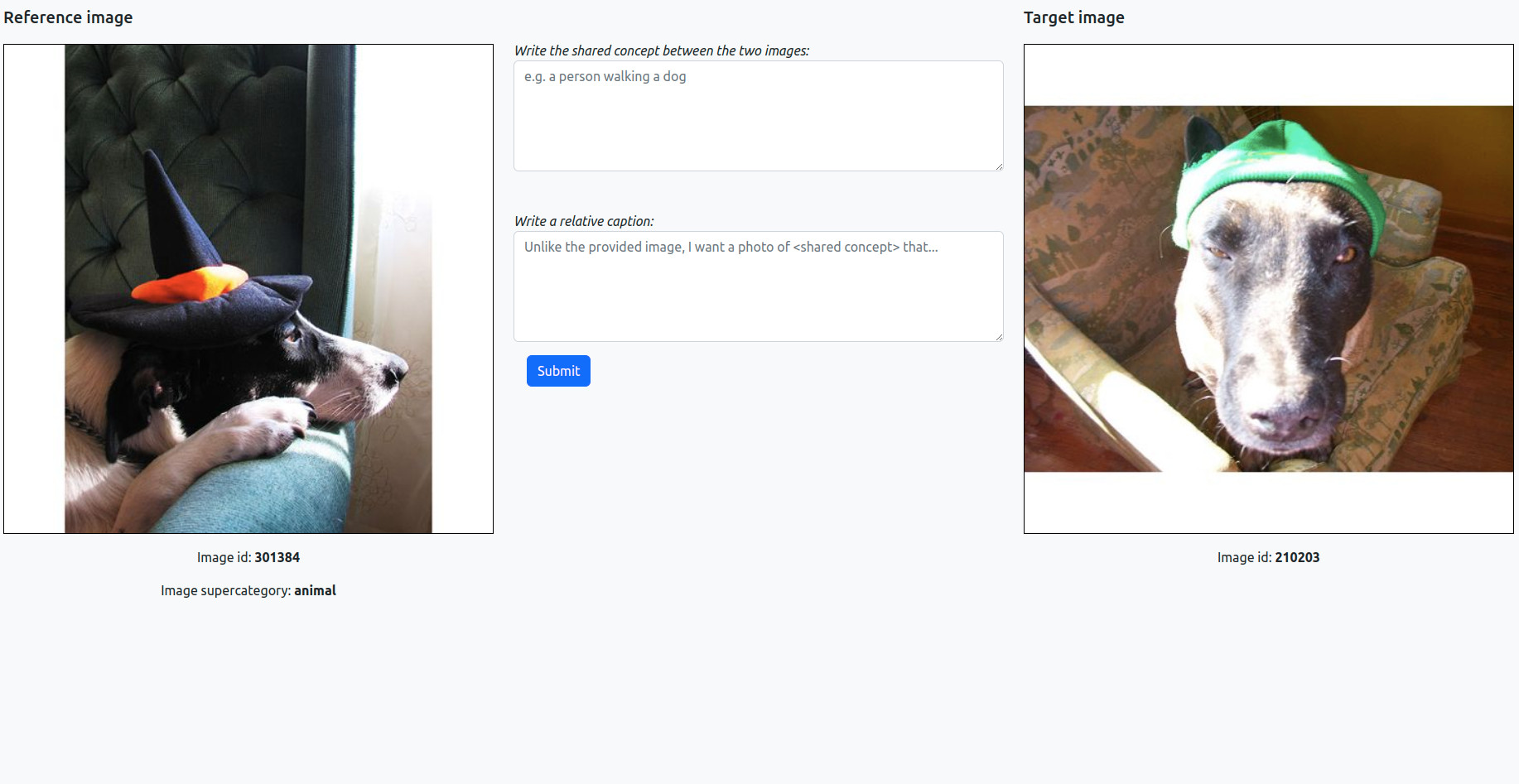}}
    \caption{Interface for writing the shared concept and the relative caption of a given reference-target image pair. \vspace{10pt}}
    \label{fig:annotator_caption}
  \end{subfigure}
  \begin{subfigure}{\linewidth}
    \frame{\includegraphics[width=\linewidth]{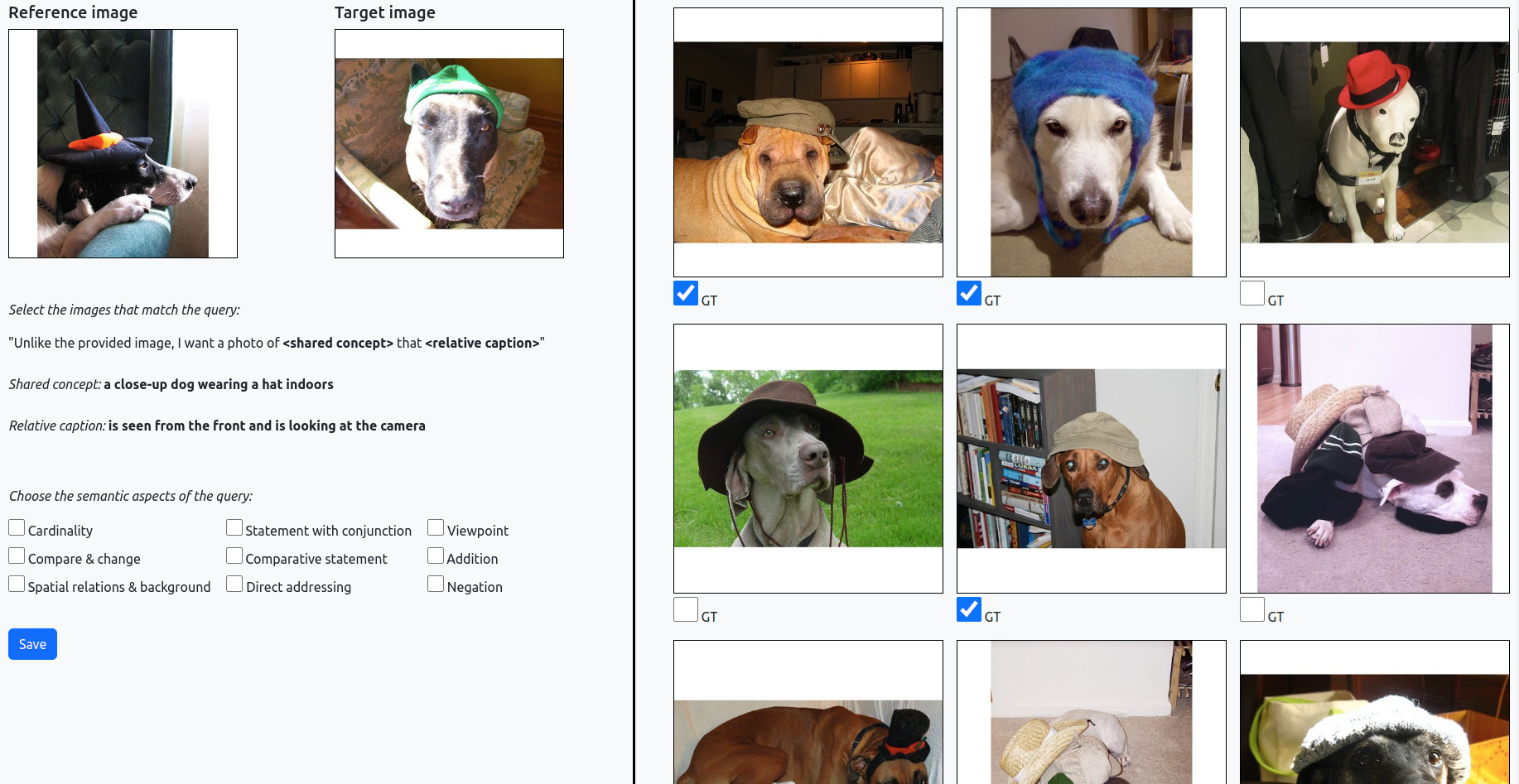}}
    \caption{Interface for selecting the multiple ground truths. On the left, the tool displays the current triplet and checkboxes to assign the semantic aspects covered by the relative caption. On the right, there is a gallery of images from which we select the ground truths.}
    \label{fig:annotator_gt_selector}
  \end{subfigure}
  \caption{Screenshots of the annotation tool user interface. (a) and (b) correspond to the first annotation phase, while (c) is related to the second one.}
  \label{fig:annotator}
\end{figure}

\section{CIRCO Dataset}
In this section, we provide details about the annotation process of the proposed CIRCO dataset, which consists of two phases. In the first phase, we build the triplets composed of a reference image, a relative caption, and a single target image. In the second one, we extend each triplet by annotating additional ground truths. The whole annotation process has been carried out by the authors of this paper. We also report a detailed analysis of CIRCO, along with a comparison with CIRR \cite{liu2021image}. 
CIRCO is available at \href{https://github.com/miccunifi/CIRCO}{\small{\url{https://github.com/miccunifi/CIRCO}}}. 

\subsection{Triplets Annotation}\label{sec:triplets_annotation}
CIRCO consists of images belonging to COCO 2017 \cite{lin2014microsoft} unlabeled set, which comprises 123,403 images. We chose this dataset as it contains open-domain real-life images depicting a large variety of subjects. We rely on the unlabeled set of COCO instead of the training one as the latter is often employed as a pre-training dataset, and we do not want any model to have a prior on the images. Every object in each image of COCO labeled sets is associated with a supercategory. The supercategories are 12 and are as follows: \textit{person}, \textit{animal}, \textit{sports}, \textit{vehicle}, \textit{food}, \textit{accessory}, \textit{electronic}, \textit{kitchen}, \textit{furniture}, \textit{indoor}, \textit{outdoor}, and \textit{appliance}. 

We start by associating every image of the unlabeled set to a supercategory relying on CLIP ViT-L/14 zero-shot classification capabilities. We assume that each image is classified according to its main subject.  Our objective is to obtain a rough estimation of the content of each image to later build a balanced dataset. Indeed, we create the queries so that we evenly distribute the reference images of CIRCO among the supercategories.
This balancing process is crucial, as we have noticed a significant domain bias within the COCO images. For instance, some objects like stop signs or fire hydrants are over-represented.

\Cref{fig:annotator_index} shows the annotation tool we employed for creating the triplets. 
The tool randomly samples a reference image and displays it next to a gallery of 50 candidate target images. 
Since CIR requires the differences between the reference and target images to be describable with a relative caption, they should be similar but with appreciable disparities. Therefore, the candidate target images are the most visually similar to the reference one according to the CLIP features. To avoid near-identical images, we filter out those with a similarity higher than 0.92. The tool allows the annotators to skip the current reference image if there is no suitable target in the gallery. Otherwise, when the annotator selects a target image, the tool displays the user interface shown in \cref{fig:annotator_caption}. In this stage, the user must write the \textit{shared concept}, \ie the shared characteristics between the reference and target images. This concept is collected to clarify possible ambiguities. For instance, the shared concept for the reference-target pair shown in \cref{fig:annotator_caption} is ``a close-up dog wearing a hat indoors". Finally, the annotator writes the relative caption from the prefix ``Unlike the provided image, I want a photo of \{shared concept\} that". To build a more challenging dataset containing truly relative captions, we ensure they do not refer to the subjects mentioned in the shared concept. Indeed, we want the subject of the relative caption to be inferred from the reference image. 

At the end of this phase, we have 1020 triplets composed of a reference image, a relative caption, and a single target image.

\subsection{Multiple Ground Truths Annotation}\label{sec:multiple_gt_annotation}
For each triplet, we want to label as ground truths all the images beside the target one that are valid matches for the corresponding query. \Cref{fig:annotator_gt_selector} shows the annotation tool we relied on in this phase. We provide the annotator with the starting triplet, and they have to select the ground truths from a gallery of images. In addition, the user also has to choose the semantic aspects of the query (see \cref{sec:dataset_analysis} for more details).

We propose leveraging our approach to ease the annotation process by employing it to retrieve the images from which we select the multiple ground truths.
Specifically, we use \method-XL to obtain the pseudo-word $S_*$ associated with the reference image. Then, we perform text-to-image retrieval with the query ``a photo of \{shared concept\} $S_*$ that \{relative caption\}". During the annotation process, we also include the shared concept in the query, as it improves performance. In fact, considering the single ground truth triplets obtained in \cref{sec:triplets_annotation}, with the shared concept we achieve a Recall$@100$ of 82.15, compared to 66.25 without it. In the gallery of images from which we select the multiple ground truths, the annotation tool provides the top 100 retrieved images with our approach and the top 50 ones most similar to the target one.

At the end of this phase, we have 4624 ground truths, of which 4097 were retrieved employing our method and 527 using the similarity with the target image. Given that with \method we achieve a Recall$@100$ of 82.15, by approximation, we can assume that in the top 100 retrieved images, there will be an average of 82.15\% of the total ground truths. This implies that the estimated number of ground truths in the entire dataset is $4097 / 0.8215 \approx 4,987$. Since we labeled 4624 images as ground truth, we can infer that the annotated ones are $4,624 / 4,987 \approx 92.7\%$ of the total. Therefore, we estimate that this annotation strategy allows us to reduce the percentage of missing ground truths in the dataset to less than 10\%.

Thanks to this second annotation phase, we labeled additional $4624\!-\!1020\!=\!3604$ ground truths that otherwise would have been false negatives. Moreover, it allows us to estimate the fraction of missing ground truths in the dataset. It is not possible to make this estimation for CIR datasets with a single ground truth, such as FashionIQ \cite{wu2021fashion} and CIRR \cite{liu2021image}, as they have not any information about the total number of ground truths. Indeed, their annotation process stops after building the triplets.

  \begin{figure}
     \centering
     \includegraphics[width=\linewidth]{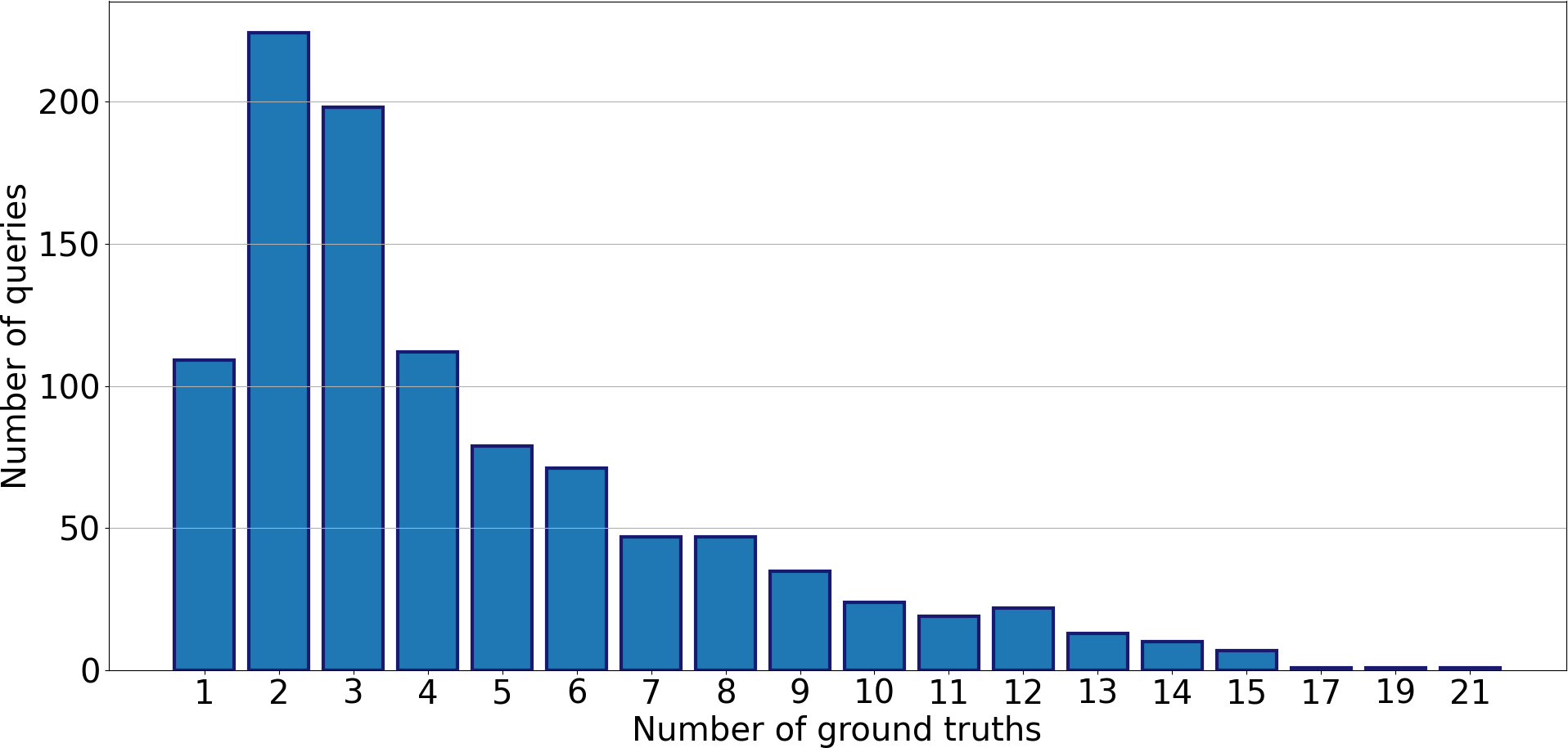}
     \caption{Histogram of the number of queries per number of ground truths for the CIRCO dataset.}
     \label{fig:gt_histogram}
 \end{figure}

\begin{figure}
     \centering
     \includegraphics[width=0.5\linewidth]{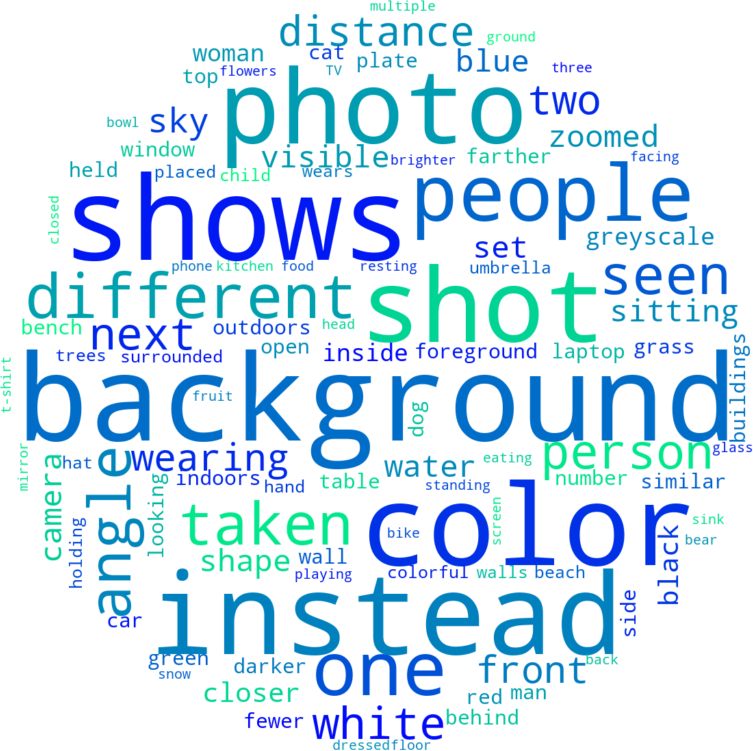}
     \caption{Vocabulary of the most frequent words in the relative captions scaled by frequency.}
     \label{fig:wordcloud}
 \end{figure}

 \begin{table}
     \centering
     \Large
     \resizebox{\linewidth}{!}{ 
     \begin{tabular}{lccc} 
       \toprule
       \multicolumn{1}{l}{\multirow{2}{*}{Semantic Aspect}} & \multicolumn{3}{c}{Coverage (\%)} \\ 
       & \multicolumn{1}{c}{CIRCO} & \multicolumn{1}{c}{CIRR} & \multicolumn{1}{c}{FashionIQ} \\
       \midrule
       Cardinality & 22.1 & 29.3$^\dagger$ & -- \\
       Addition & 26.7 & 15.2$^\dagger$ & 15.7$^\dagger$ \\
       Negation & 12.2 & 11.9$^\dagger$ & ~4.0$^\dagger$ \\ 
       Direct Addressing & 50.7 & 57.4$^\dagger$ & 49.0$^\dagger$ \\
       Compare \& Change & 32.0 & 31.7$^\dagger$ & ~3.0$^\dagger$ \\
       Comparative Statement & 35.3 & 51.7$^\dagger$ & 32.0$^\dagger$ \\
       Statement with Conjunction & 75.1 & 43.7$^\dagger$ & 19.0$^\dagger$ \\ 
       Spatial Relations \& Background & 45.7 & 61.4$^\dagger$ & -- \\
       Viewpoint & 22.6 & 12.7$^\dagger$ & -- \\
       \midrule
       \textit{Avg. Caption Length (words)} & 10.4 & 11.3$^\dagger$ & 5.3$^\dagger$ \\
       \bottomrule
       \end{tabular}}
   \caption{Analysis of the semantic aspects covered by the relative captions. $^{\dagger}$ indicates results taken from \cite{liu2021image}. -- denotes no reported results.}
   \label{tab:caption_semantic}
 \end{table}

  \begin{figure*}
  \centering
  \captionsetup[subfigure]{aboveskip=2pt}
  \begin{subfigure}{0.3\linewidth}
    \includegraphics[width=\linewidth]{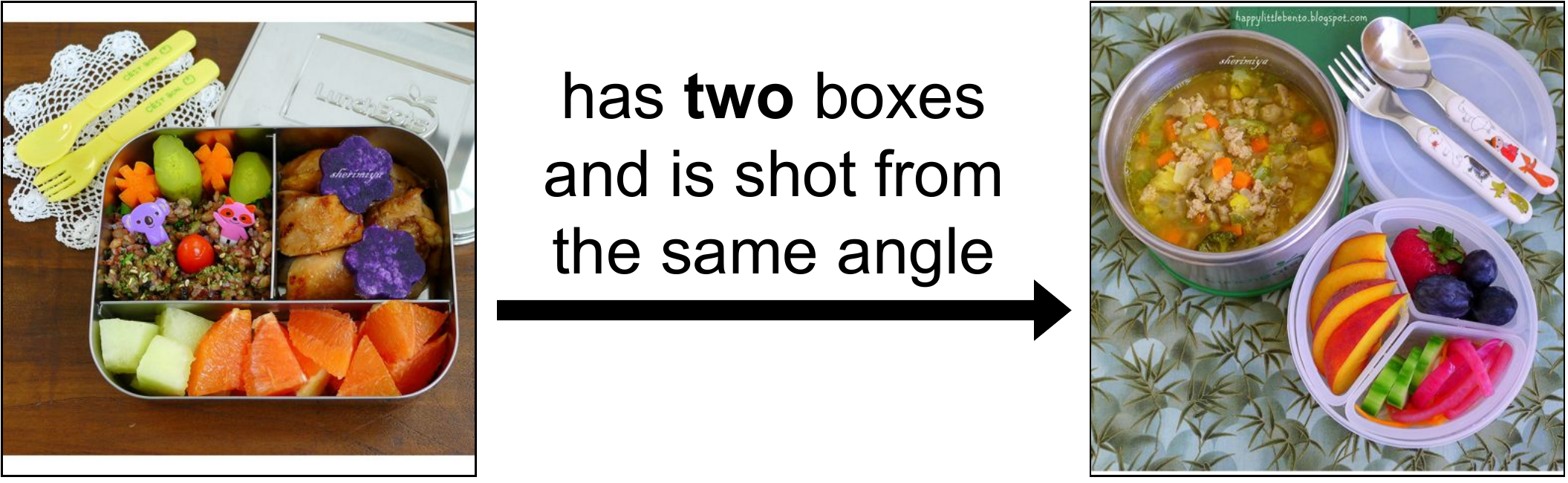}
    \caption{Cardinality \vspace{10pt}}
    \label{fig:circo_example_cardinality}
  \end{subfigure}
  \hfill
  \begin{subfigure}{0.3\linewidth}
    \includegraphics[width=\linewidth]{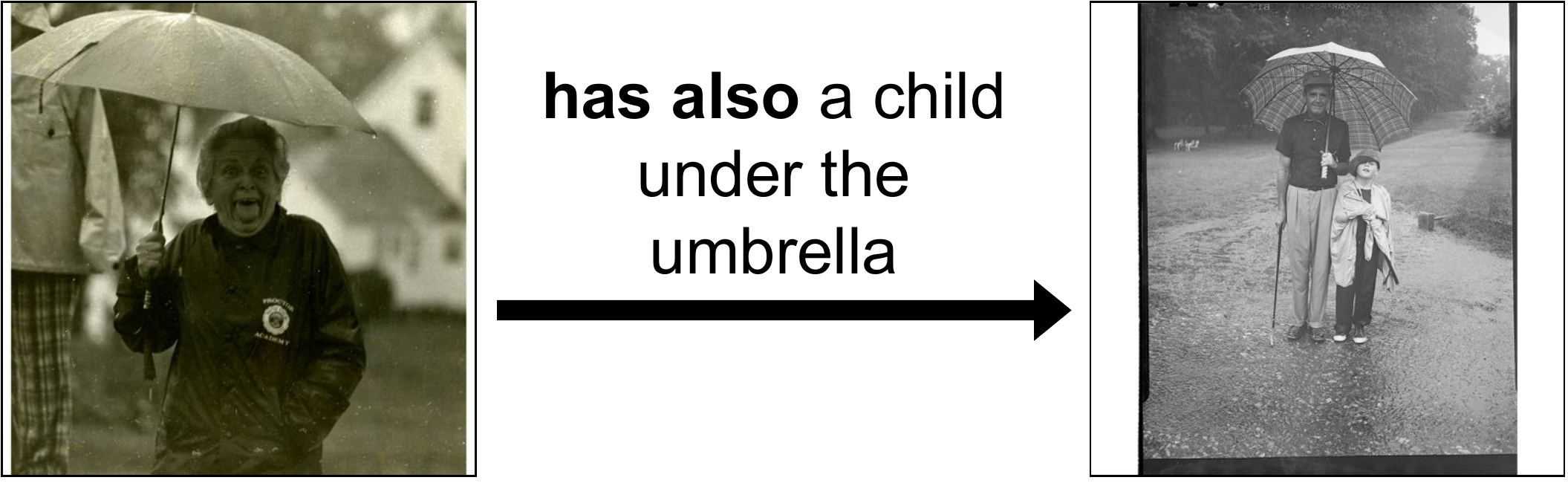}
    \caption{Addition \vspace{10pt}}
    \label{fig:circo_example_addition}
  \end{subfigure}
  \hfill
  \begin{subfigure}{0.3\linewidth}
    \includegraphics[width=\linewidth]{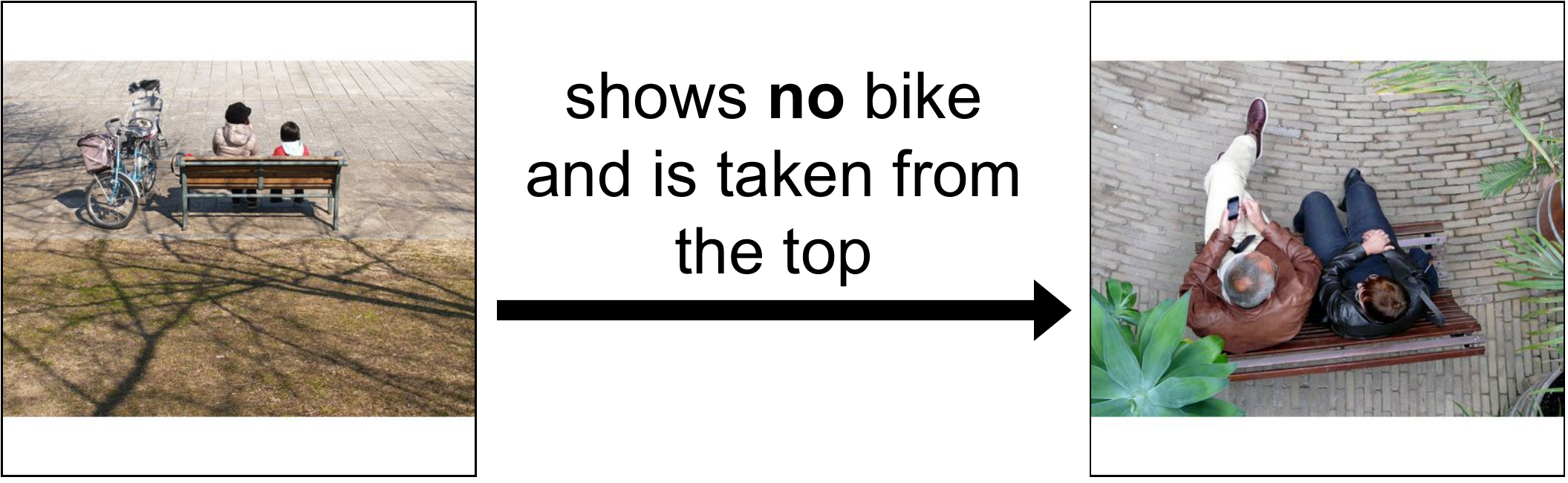}
    \caption{Negation \vspace{10pt}}
    \label{fig:circo_example_negation}
  \end{subfigure}
  \vspace{10pt}
  \begin{subfigure}{0.3\linewidth}
    \includegraphics[width=\linewidth]{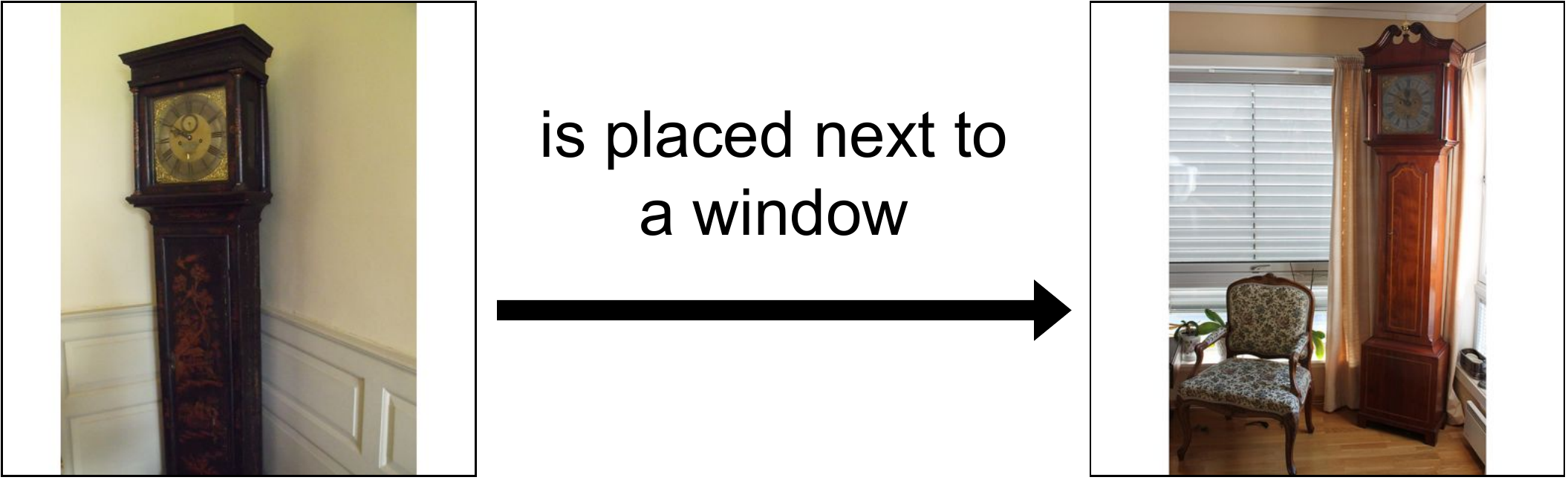}
    \caption{Direct Addressing}
    \label{fig:circo_example_direct_addressing}
  \end{subfigure}
  \hfill
  \begin{subfigure}{0.3\linewidth}
    \includegraphics[width=\linewidth]{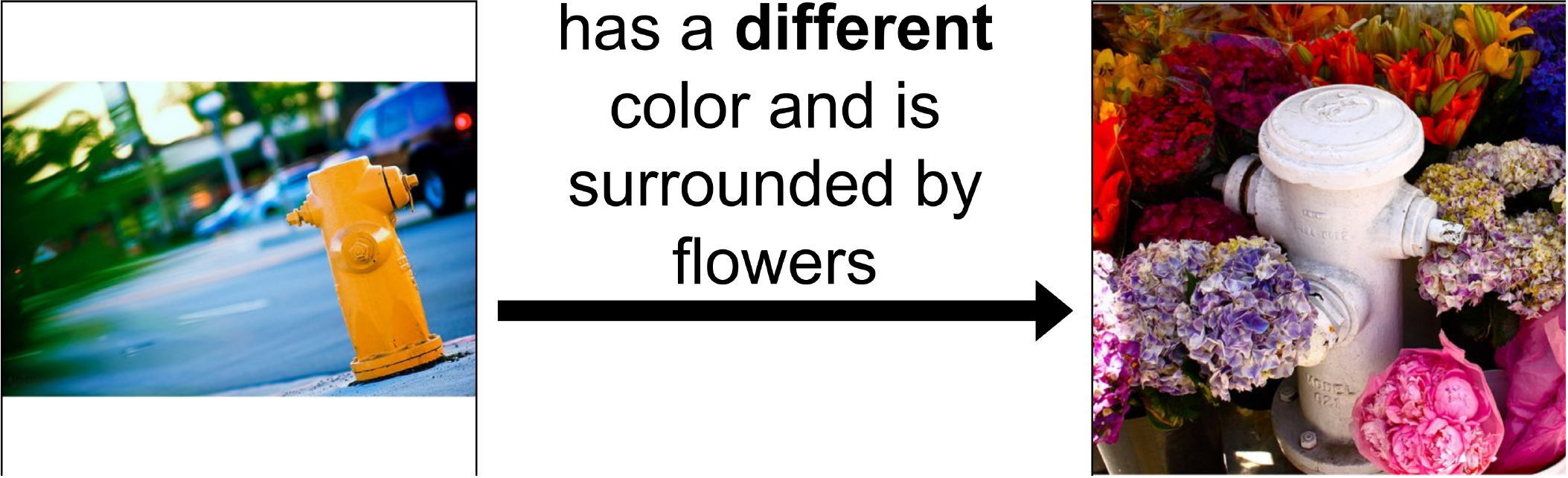}
    \caption{Compare \& Change}
    \label{fig:circo_example_compare_change}
  \end{subfigure}
  \hfill
  \begin{subfigure}{0.3\linewidth}
    \includegraphics[width=\linewidth]{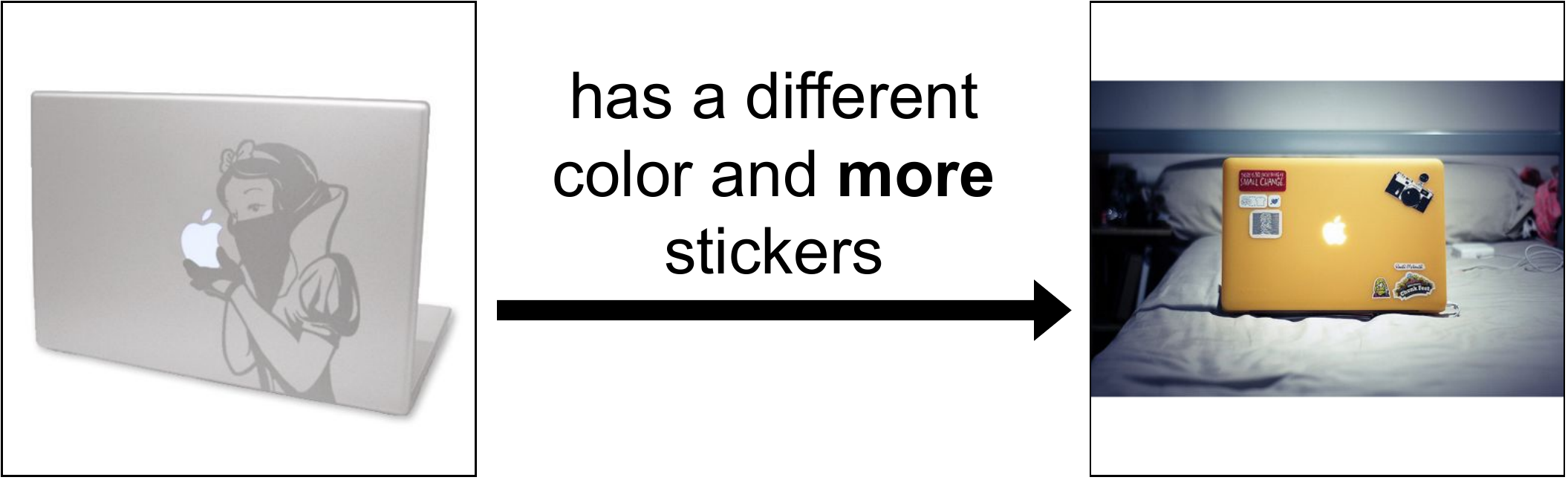}
    \caption{Comparative Statement}
    \label{fig:circo_example_comparative_statement}
  \end{subfigure}
  \begin{subfigure}{0.3\linewidth}
    \includegraphics[width=\linewidth]{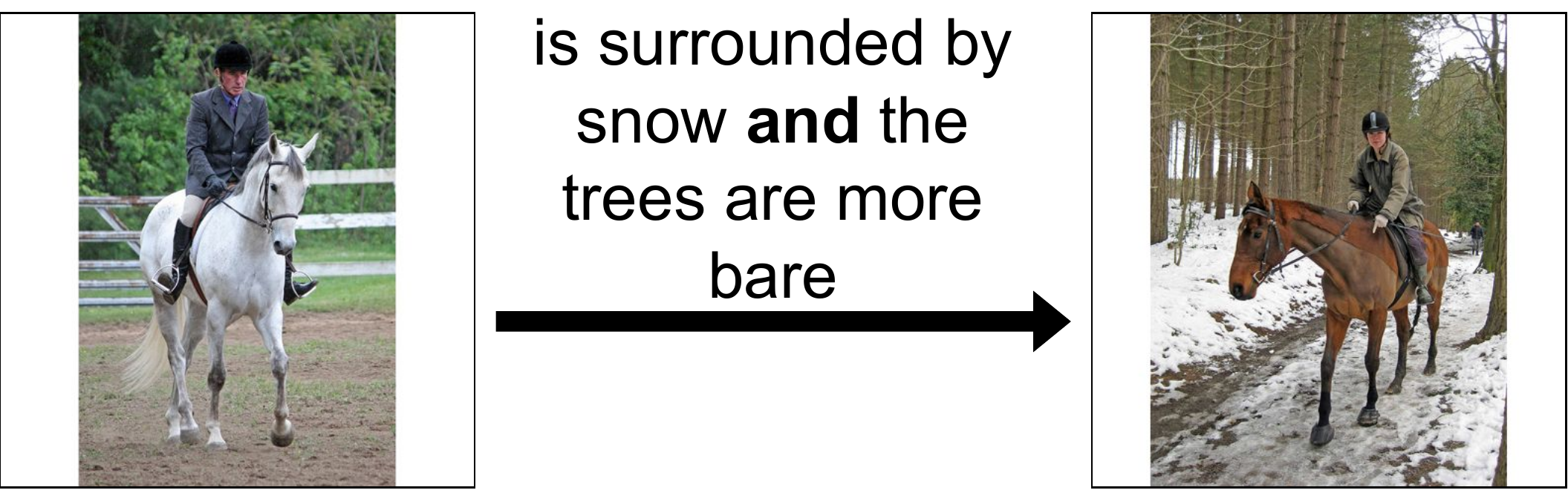}
    \caption{Statement with Conjunction}
    \label{fig:circo_example_statement_conjunction}
  \end{subfigure}
  \hfill
  \begin{subfigure}{0.3\linewidth}
    \includegraphics[width=\linewidth]{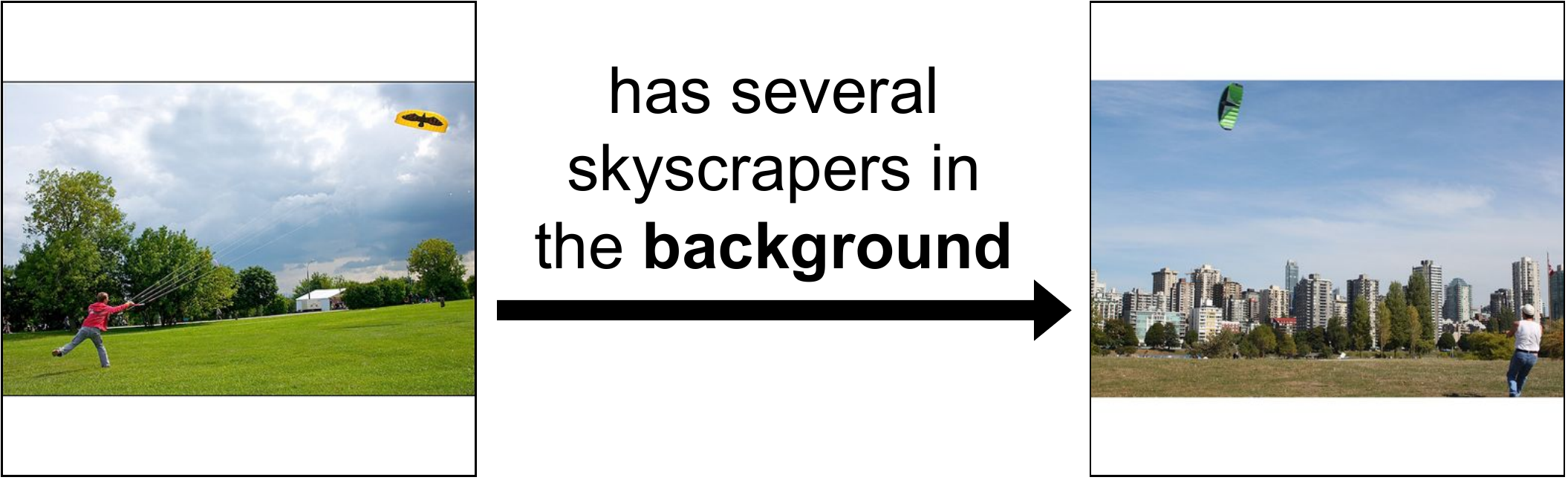}
    \caption{Spatial Relations \& Background}
    \label{fig:circo_example_background}
  \end{subfigure}
  \hfill
  \begin{subfigure}{0.3\linewidth}
    \includegraphics[width=\linewidth]{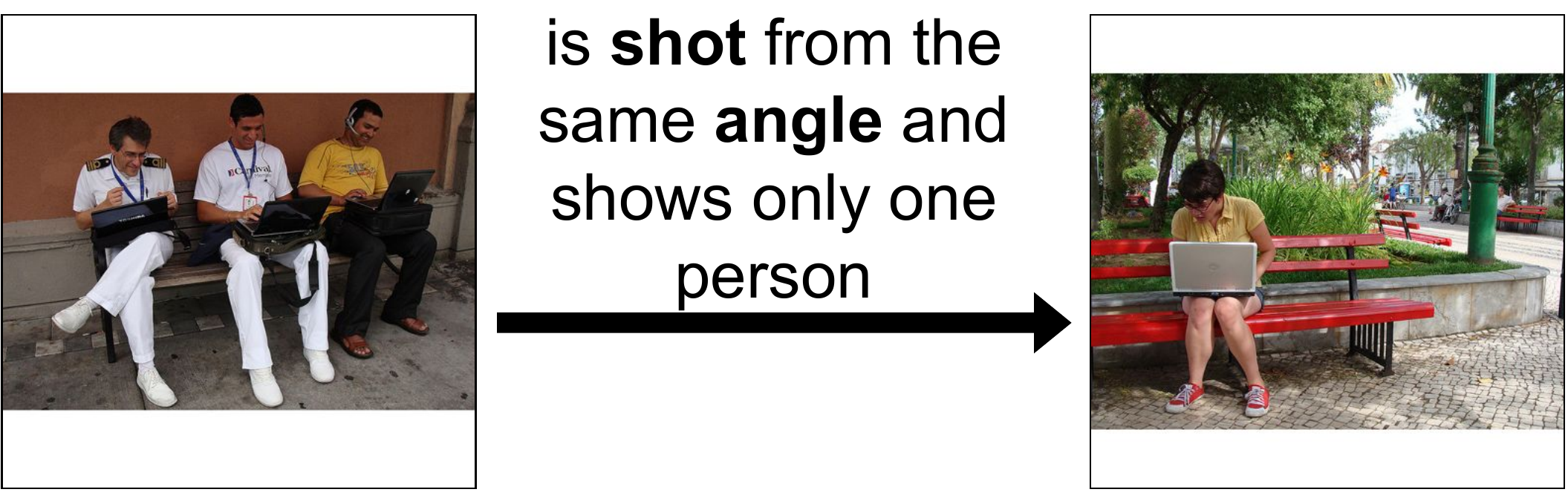}
    \caption{Viewpoint}
    \label{fig:circo_example_viewpoint}
  \end{subfigure}
  \caption{Examples of queries of the proposed CIRCO dataset for different semantic aspects. For simplicity, we report only one ground truth. We highlight the keywords of each semantic aspect in bold.}
  \label{fig:circo_semantic_examples}
\end{figure*}
\begin{figure*}
     \centering
     \includegraphics[width=\linewidth]{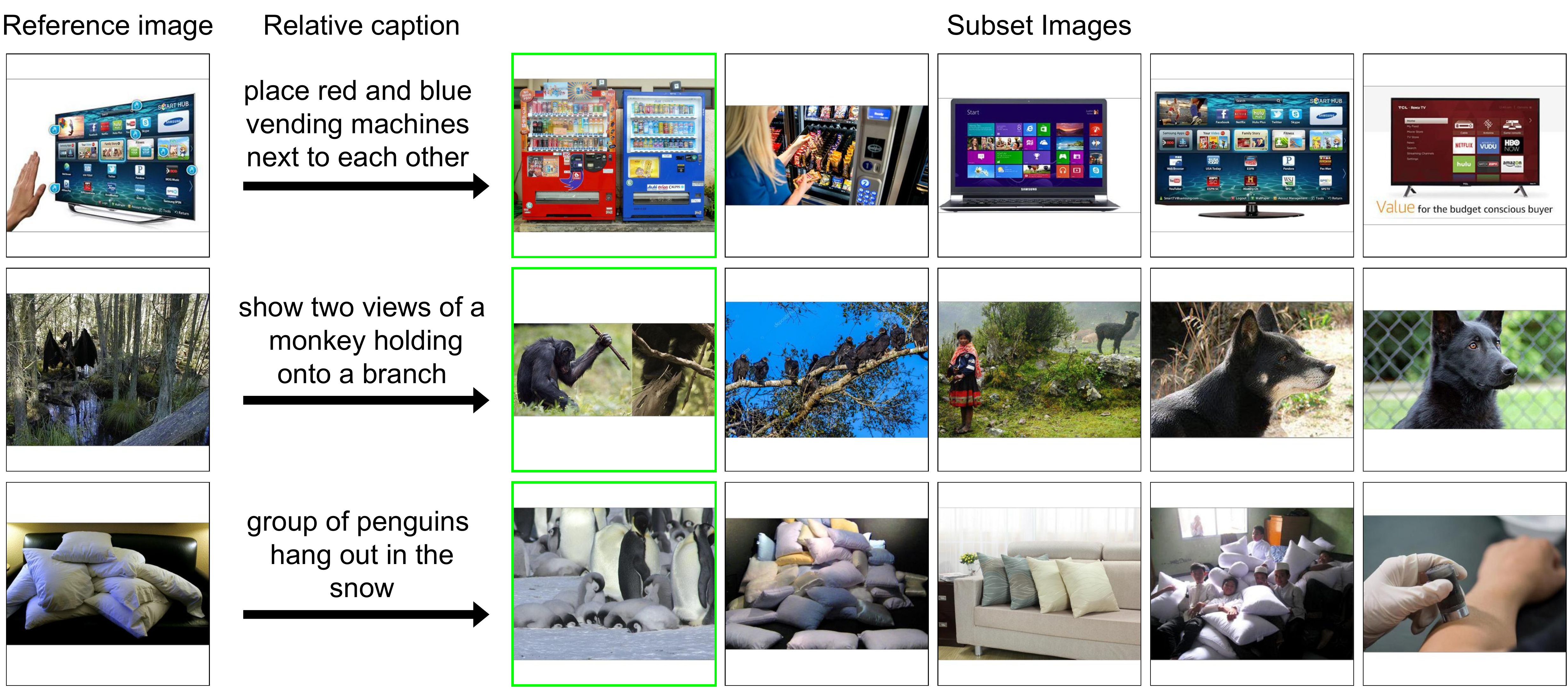}
     \caption{Examples of queries belonging to the CIRR dataset \cite{liu2021image}. The subsets of images depict very different subjects and the relative captions do not consider the reference images. We highlight the target image with a green border.}
     \label{fig:cirr_subset_example}
 \end{figure*}

\subsection{Dataset Analysis} \label{sec:dataset_analysis}
CIRCO comprises 1020 queries with a total of 4624 ground truths, 4.53 per query on average. \Cref{fig:gt_histogram} shows a histogram representing the number of queries per number of ground truths. The maximum number of ground truths annotated for a query is 21, while the modal value is 2. 

The relative captions are composed of an average of 10.4 words.
\Cref{fig:wordcloud} illustrates a word cloud of the most frequent words in the annotations. Following CIRR \cite{liu2021image}, we analyze the semantic aspects covered by the relative captions. \Cref{tab:caption_semantic} reports the results. We observe that the average length of the captions and the distribution of the semantic concepts is comparable with CIRR. However, in CIRCO about 75\% of the annotations are composed of multiple statements, more than the $\sim$43\% of CIRR, thus revealing a higher complexity. We provide a query example for each semantic aspect in \cref{fig:circo_semantic_examples}.

CIRR \cite{liu2021image} validation and test sets contain 4K triplets each. We recall that during the data collection process, CIRR constructs subsets of 6 visually similar images in an automated way according to the features of a ResNet152 \cite{he2016deep}. Then, the queries are built such that the reference and the target images belong to the same subset. However, despite the feature similarity, the images of the subset often depict very different subjects. This makes writing a relative caption unfeasible for a human annotator and leads to an absolute description of the target image. We report some examples of this issue in \cref{fig:cirr_subset_example}. We observe that, for instance, the annotator needs to rely on an absolute caption to describe the differences between an image depicting some pillows and one with a group of penguins.

CIRCO annotation strategy aims to address the issue mentioned above. Indeed, we let the annotators choose the reference-target pair without any constraint. This way, we ensure that the annotators only write captions that are actually relative, thereby increasing the quality of the dataset.

CIRCO comprises 1020 queries, randomly divided into 220 and 800 for the validation and test set, respectively. Compared to CIRR, we have fewer queries, but our two-phase annotation strategy ensures higher quality, reduced false negatives, and the availability of multiple ground truths. Moreover, we provide significantly more distractors than the 2K images of the CIRR test set by employing all the 120K images of COCO as the index set. \Cref{fig:circo_semantic_examples} shows some query examples.

\subsection{Dataset Evaluation}
Thanks to the reduced false negatives and multiple ground truths, for performance evaluation on CIRCO we adopt the fine-grained metric mean Average Precision (mAP). In particular, we compute mAP$@K$, with K ranging from 5 to 50, as follows:
\begin{equation}
    \text{mAP}@K = \frac{1}{N} \sum\limits^N_{n=1} \frac{1}{\min(K, G_n)} \sum\limits_{k=1}^K P@k * \text{rel}@k
\end{equation}
where $N$ is the number of queries, $G_n$ is the number of ground truths of the $n$-th query, $P@k$ is the precision at rank $k$, rel$@k$ is a relevance function. The relevance function is an indicator function that equals 1 if the image at rank $k$ is labeled as positive and equals 0 otherwise.

 \section{Additional Experimental Results}

 \begin{table*}[!ht]
  \centering
  \Large
  \resizebox{0.6\linewidth}{!}{ 
  \begin{tabular}{clcccccccc} 
  \toprule
  \multicolumn{1}{c}{} & \multicolumn{1}{c}{} & \multicolumn{3}{c}{IR} & \multicolumn{1}{c}{} & \multicolumn{4}{c}{CIR}\\
  \cmidrule(lr){3-5}\cmidrule(lr){7-10}
  \multicolumn{1}{l}{Ablation} & \multicolumn{1}{l}{Method} & R$@1$ & R$@3$ & R$@5$ &  & R$@1$ & R$@5$ & R$@10$ & R$@50$  \\ \midrule
  \multirow{4}{*}{OTI} & w/o GPT reg & 99.58 & 99.90 & 99.96 &  & \underline{21.63} & \underline{50.51} & \underline{64.07} & \underline{88.04} \\
  & random reg & 99.03 & 99.21 & 99.55 &  & 21.09 & 50.42 & 63.84 & 87.30\\
  & w/o reg & \underline{99.72} & \underline{99.91} & \textbf{100} &  & 19.30 & 46.81 & 59.96 &  84.74\\
  & \textbf{\method-OTI} & \textbf{99.81} & \textbf{100} & \textbf{100} &  & \textbf{23.54} & \textbf{53.93} & \textbf{67.69} & \textbf{90.31}\\ \midrule[.02em]
  \multirow{2}{*}{$\phi$} & w/o reg & \textbf{99.35} & \textbf{100} & \textbf{100} &  & \underline{22.41} & \underline{53.00} & \underline{66.90} & \underline{89.95}\\
   & \textbf{\method} & \underline{98.66} & \underline{99.86} & \textbf{100} &  & \textbf{25.09}  & \textbf{55.18} & \textbf{68.79} & \textbf{90.82}\\
  \bottomrule
  \end{tabular}}
  \caption{Evaluation of the visual information embedded in $v_*$ for different regularization techniques on CIRR validation set. IR and CIR stand for Image Retrieval and Composed Image Retrieval, respectively. Best and second-best scores are highlighted in bold and underlined, respectively.}
  \label{tab:vstar_image_retrieval}
\end{table*}

\begin{table*}[!t]
  \centering  
  \resizebox{0.8\linewidth}{!}{ 
  \begin{tabular}{clcccccccc} 
  \toprule
  \multicolumn{1}{c}{} & \multicolumn{1}{l}{} & \multicolumn{2}{c}{Shirt} & \multicolumn{2}{c}{Dress} &\multicolumn{2}{c}{Toptee} &\multicolumn{2}{c}{Average} \\
  \cmidrule(lr){3-4}
  \cmidrule(lr){5-6}
  \cmidrule(lr){7-8}
  \cmidrule(lr){9-10}
  \multicolumn{1}{c}{Backbone} & \multicolumn{1}{l}{Method} & R$@10$ & R$@50$ & R$@10$ & R$@50$ & R$@10$ & R$@50$ & R$@10$ & R$@50$ \\ 
  \midrule
  \multirow{4}{*}{B/32} & \method-FIQ & \textbf{26.15} & \textbf{43.57} & \textbf{20.62} & \textbf{42.69} & \textbf{27.89} & \textbf{49.36} & \textbf{24.89} & \textbf{45.21} \\
  & \method-CIRR & 24.44 & 41.24 & 18.29 & 38.92 & 25.40 & 45.69 & 22.71 & 41.95 \\
   & \textbf{\method} & 24.44 & \underline{41.61} & \underline{18.54} & 39.51 & \underline{25.70} & \underline{46.46} & \underline{22.89} & \underline{42.53}  \\ 
   & \textbf{\method-OTI} & \underline{25.37} & 41.32 & 17.85 & \underline{39.91} & 24.12 & 45.79 & 22.44 & 42.34
  \\ \midrule[.02em]
 
    \multirow{4}{*}{L/14} & \method-XL-FIQ & \underline{29.54} & \textbf{48.04} & \textbf{23.15} & \textbf{46.36} & \textbf{31.16} & \textbf{53.34} & \textbf{27.95} & \textbf{49.24}  \\

   & \method-XL-CIRR & 25.22 & 42.44 & 19.29 & 41.00 & 27.38 & 48.29 & 23.96 & 43.91 \\
  & \textbf{\method-XL} & 26.89 & 45.58 & 20.48 & 43.13 & 29.32 & 49.97 & 25.56 & 46.23 \\ 
   & \textbf{\method-XL-OTI} & \textbf{30.37} & \underline{47.49} & \underline{21.57} & \underline{44.47} & \underline{30.90} & \underline{51.76} & \underline{27.61} & \underline{47.90} \\
  \bottomrule
  \end{tabular}}
  \caption{Quantitative results of our approach on FashionIQ validation set varying the pre-training dataset of $\phi$. Best and second-best scores are highlighted in bold and underlined, respectively.}
  \label{tab:fashioniq_phi_ablation}
\end{table*}

\subsection{Visual Information in $\vect{v_*}$} 
To evaluate the effectiveness of the pseudo-word tokens in capturing visual information, we conduct an image retrieval experiment. Specifically, we investigate whether the pseudo-word tokens are able to retrieve the corresponding images. 

Given an input image $I$, we perform textual inversion to obtain the corresponding pseudo-word token $v_*$ and its associated pseudo-word $S_*$. We build a generic prompt using the pseudo-word $S_*$ such as ``a photo of $S_*$". We extract the text features using CLIP text encoder $\psi_{T}$ and use them to query an image database. If the pseudo-word token manages to capture the visual content of the input image, we expect the image $I$ to be the top-ranked result.

\Cref{tab:vstar_image_retrieval} shows the results for Image Retrieval (IR) next to the corresponding ones for Composed Image Retrieval (CIR). We carry out all the experiments on the CIRR validation set. We report the results obtained by all the ablation studies on the regularization loss for both OTI and $\phi$. We observe that, regardless of the regularization technique, $v_*$ captures the visual information of the image effectively. However, we notice a significant improvement in the performance of CIR when using the proposed GPT-powered loss. This proves how our regularization technique enhances the ability of the pseudo-word tokens to interact with the actual words that compose the relative caption.

\subsection{Training $\vect{\phi}$ on Different Datasets}
We perform several experiments to investigate the impact of the pre-training dataset we employed for training the textual inversion network $\phi$. In particular, besides the version of $\phi$ trained on the test split of ImageNet1K, we also train two variants using the training sets of CIRR and FashionIQ, named \method-CIRR and \method-FIQ, respectively. Notably, we rely only on the raw images of these datasets without considering the associated labels. This way, our approach is still unsupervised. The FashionIQ and CIRR training sets comprise 45,429 and 16,939 images, respectively. For both datasets, the number of images is lower than the 100K contained in the ImageNet1K test split. To assess the generalization capabilities of our approach, we test all three variants of the $\phi$ network on both the FashionIQ and CIRR validation sets. 

\Cref{tab:fashioniq_phi_ablation} shows the results on the FashionIQ validation set. We notice how \method-FIQ and \method-XL-FIQ improve the performance over the ImageNet-based variants. We suppose this gain is related to the narrow domain of FashionIQ, which has a much more limited scope than the natural images of ImageNet. Furthermore, both \method-FIQ and \method-XL-FIQ outperform the OTI-based methods, showing the effectiveness of our distillation-based approach. 
Regarding the CIRR variant of $\phi$, we observe that with the B/32 backbone, it achieves comparable performance to \method, while with the L/14 one, the results are worse than those of \method-XL but still noteworthy. Considering that the CIRR training set consists of only 16K images, we can infer that our approach is effective even in a low-data regime. 

\begin{table*}[!t]
  \centering
  \resizebox{0.8\linewidth}{!}{ 
  \begin{tabular}{clccccccc} 
  \toprule
  \multicolumn{1}{c}{} & \multicolumn{1}{c}{} & \multicolumn{4}{c}{Recall$@K$} & \multicolumn{3}{c}{Recall$_{\text{Subset}}@K$} \\
  \cmidrule(lr){3-6}
  \cmidrule(lr){7-9}
  \multicolumn{1}{l}{Backbone} & \multicolumn{1}{l}{Method} & $K = 1$ & $K = 5$ & $K = 10$ & $K = 50$ & $K = 1$ & $K = 2$ & $K = 3$ \\ 
  \midrule
  \multirow{4}{*}{B/32} & \method-FIQ & 23.99 & 53.53 & 67.33 & 89.17 & \textbf{57.07} & \textbf{78.21} & \textbf{89.28} \\ 
   & \method-CIRR & \textbf{25.28} & \textbf{55.32} & \underline{68.74} & \textbf{90.89} & \underline{55.18} & 76.27 & \underline{88.07} \\
   & \textbf{\method} & \underline{25.09}  & \underline{55.18} & \textbf{68.79} & \underline{90.82} & 54.84 & \underline{76.63} & 87.95  \\
   & \textbf{\method-OTI} & 23.54 & 53.93 & 67.69 & 90.31 & 51.26 & 73.02 & 86.51  \\
   \midrule[.02em]
    \multirow{4}{*}{L/14} & \method-XL-FIQ & 24.04 & 53.67 & 66.92 & 88.07 & \textbf{55.80} & \textbf{77.13} & \textbf{88.26} \\ 
   & \method-XL-CIRR & \textbf{24.61} & 53.79 & \underline{67.06} & 88.85 & \underline{54.39} & \underline{75.68} & 87.37 \\
  & \textbf{\method-XL} & 24.11 & \underline{54.25} & 66.95 & \textbf{89.48} & 53.77 & 75.29 & \underline{87.56} \\
   & \textbf{\method-XL-OTI} & \underline{24.40} & \textbf{54.68} & \textbf{68.02} & \underline{89.09} & 52.27 & 74.53 & 86.80  \\
  \bottomrule
  \end{tabular}}
  \caption{Quantitative results of our approach on CIRR validation set varying the pre-training dataset of $\phi$. Best and second-best scores are highlighted in bold and underlined, respectively.}
  \label{tab:cirr_phi_ablation}
\end{table*}

In \cref{tab:cirr_phi_ablation}, we report the results on the CIRR validation set. Interestingly, we notice that \method-FIQ and \method-XL-FIQ manage to generalize to a broader domain achieving promising performance. In addition, due to the domain similarity between the CIRR and ImageNet datasets, we observe that the CIRR-based versions of $\phi$ obtain comparable results to the ImageNet-based ones.

\subsection{CIRCO}
\Cref{tab:circo_val} shows the results on the CIRCO validation set. We observe that the same considerations made for the test set \cref{tab:circo_test} in \cref{sec:experimental_results} hold true. We still report these results for completeness.

Additionally, we evaluate the performance of \method on the CIRCO test set considering only the first annotated ground truth (end of \cref{sec:triplets_annotation}) in \cref{tab:circo_single_gt}. Since we leverage  \method-XL only during the second annotation phase, we can fairly compare with the baselines. We employ Recall$@K$ as the evaluation metric. We observe that even with a single annotated ground truth, the Image + Text baseline outperforms Image-only and Text-only. This confirms that we need both the reference image and the relative caption to retrieve the target image. 
\method and \method-XL achieve the best performance improving over all the other methods and baselines. In particular, we notice that \method-XL significantly outperforms Pic2Word while leveraging the same CLIP backbone.

\begin{table}[!ht]
  \centering
  \Large
  \resizebox{\linewidth}{!}{ 
  \begin{tabular}{clcccc} 
  \toprule
  \multicolumn{1}{c}{} & \multicolumn{1}{c}{} & \multicolumn{4}{c}{mAP$@K$} \\
  \cmidrule(lr){3-6}
  \multicolumn{1}{l}{Backbone} & \multicolumn{1}{l}{Method} & $K=5$ & $K=10$ & $K=25$ & $K=50$ \\ \midrule
  \multirow{7}{*}{B/32} & Image-only & 1.61 & 2.16 & 2.73 & 3.10 \\
  & Text-only & 2.96 & 3.29 & 3.74 & 3.89 \\
  & Image + Text & 2.63 & 3.58 & 4.52 & 4.94 \\
  & Captioning & 5.12 & 5.31 & 6.38 & 6.77 \\
  & PALAVRA \cite{cohen2022this} & 5.15 & 6.13 & 7.20 & 7.78 \\
  & \textbf{\method-OTI} & \underline{6.61} & \underline{7.24} & \underline{8.30} & \underline{8.97} \\
  & \textbf{\method} &  \textbf{6.82} & \textbf{7.83}  & \textbf{9.15} & \textbf{9.77} \\ \midrule[.02em]
  \multirow{3}{*}{L/14} & Pic2Word~\cite{saito2023pic2word} & 7.92 & 9.02 & 10.18 & 10.83\\ 
  & \textbf{\method-XL-OTI} & \textbf{10.85} & \textbf{12.15} & \textbf{13.63} & \textbf{14.46} \\
  & \textbf{\method-XL} & \underline{10.09} & \underline{11.15} & \underline{12.83} & \underline{13.60} \\
  \bottomrule
  \end{tabular}}
  \caption{Quantitative results on CIRCO validation set. Best and second-best scores are highlighted in bold and underlined, respectively.}
  \label{tab:circo_val}
\end{table}

\begin{table}[!ht]
  \centering
  \Large
  \resizebox{\linewidth}{!}{ 
  \begin{tabular}{clcccc} 
  \toprule
  \multicolumn{1}{c}{} & \multicolumn{1}{c}{} & \multicolumn{4}{c}{Recall$@K$} \\
  \cmidrule(lr){3-6}
  \multicolumn{1}{l}{Backbone} & \multicolumn{1}{l}{Method} & $K=5$ & $K=10$ & $K=25$ & $K=50$ \\ \midrule
  \multirow{7}{*}{B/32} & Image-only & 3.88 & 6.63 & 14.13 & 22.00 \\
  & Text-only & 4.75 & 6.63 & 9.50 & 13.50 \\
  & Image + Text & 8.25 & 14.13 & 25.50 & 34.75 \\
  & Captioning & 10.25 & 14.33 & 21.38 & 29.00 \\
  & PALAVRA \cite{cohen2022this} & 12.63 & 20.63 & 32.00 & 41.75 \\
  & \textbf{\method-OTI} & \underline{16.88} & \underline{25.00} & \underline{37.00} & \underline{46.38} \\
  & \textbf{\method} & \textbf{19.75} & \textbf{28.00} &  \textbf{39.50} & \textbf{50.63} \\
   \midrule[.02em]
  \multirow{3}{*}{L/14} & Pic2Word~\cite{saito2023pic2word} & 16.13 & 24.38 & 37.25 & 46.50\\
  & \textbf{\method-XL-OTI} & \underline{22.75} & \underline{32.00} & \underline{45.13} & \textbf{58.00} \\
  & \textbf{\method-XL} & \textbf{23.50} & \textbf{32.63} & \textbf{45.25} & \underline{55.63} \\
  \bottomrule
  \end{tabular}}
  \caption{Quantitative results on CIRCO test set considering only the first annotated ground truth. Best and second-best scores are highlighted in bold and underlined, respectively.}
  \label{tab:circo_single_gt}
\end{table}

\subsection{Comparison with Supervised Baselines}
We compare our zero-shot approach to a supervised method. Specifically, we consider Combiner \cite{baldrati2022effective}, which integrates image and text CLIP features using a combiner network. Since we also rely on an out-of-the-box CLIP model, we believe Combiner constitutes the most similar method to ours among the supervised ones. We train Combiner both on FashionIQ and CIRR training sets with the official repository using the B/32 backbone. To evaluate the generalization capabilities of supervised models, we test both Combiner versions on FashionIQ and CIRR validation sets and compare them with our zero-shot method. We report the results in \cref{tab:supervised_baselines}. As expected, when the training and testing datasets correspond, Combiner achieves the best results. However, we observe that both the supervised models struggle to generalize to different domains, as also noticed by \cite{saito2023pic2word}. On the contrary, \method obtains noteworthy performance on both datasets in a zero-shot manner. Therefore, as we do not require an expensive manually-annotated training set, our approach proves to be more scalable and more suitable for the broad applicability of CIR.

\begin{table}[!ht]
  \centering
  \resizebox{\linewidth}{!}{ 
  \begin{tabular}{lcccccc} 
  \toprule
\multicolumn{1}{c}{} & \multicolumn{4}{c}{CIRR} & \multicolumn{2}{c}{FashionIQ}\\
  \cmidrule(lr){2-5}
  \cmidrule{6-7}
  \multicolumn{1}{l}{Method} & R$@1$ & R$@5$ & R$@10$ & R$@50$ & R$@10$ & R$@50$ \\ \midrule
  Combiner-FIQ~\cite{baldrati2022effective} & 19.88 & 48.05 & 61.11 & 85.51 & \textbf{32.96} & \textbf{54.55} \\
  Combiner-CIRR~\cite{baldrati2022effective} & \textbf{32.24} & \textbf{65.46} & \textbf{78.21} & \textbf{95.19} & 20.91 & 40.40 \\
  \textbf{\method-OTI} & 23.54 & 53.93 & 67.69 & 90.31 & 22.44 & 42.34 \\
  \textbf{\method} & \underline{25.09} & \underline{55.18} & \underline{68.79} & \underline{90.82} & \underline{22.89} & \underline{42.53} \\  
  \bottomrule
  \end{tabular}}
  \caption{Comparison with supervised baselines on CIRR and FashionIQ validation sets. Combiner-FIQ and Combiner-CIRR denote the models from \cite{baldrati2022effective} trained on FashionIQ and CIRR, respectively. For FashionIQ, we consider the average recall. Best and second-best scores are highlighted in bold and underlined, respectively.}
  \label{tab:supervised_baselines}
\end{table}

\subsection{Qualitative Results}
\begin{figure*}
    \centering
    \includegraphics[width=0.97\textwidth]{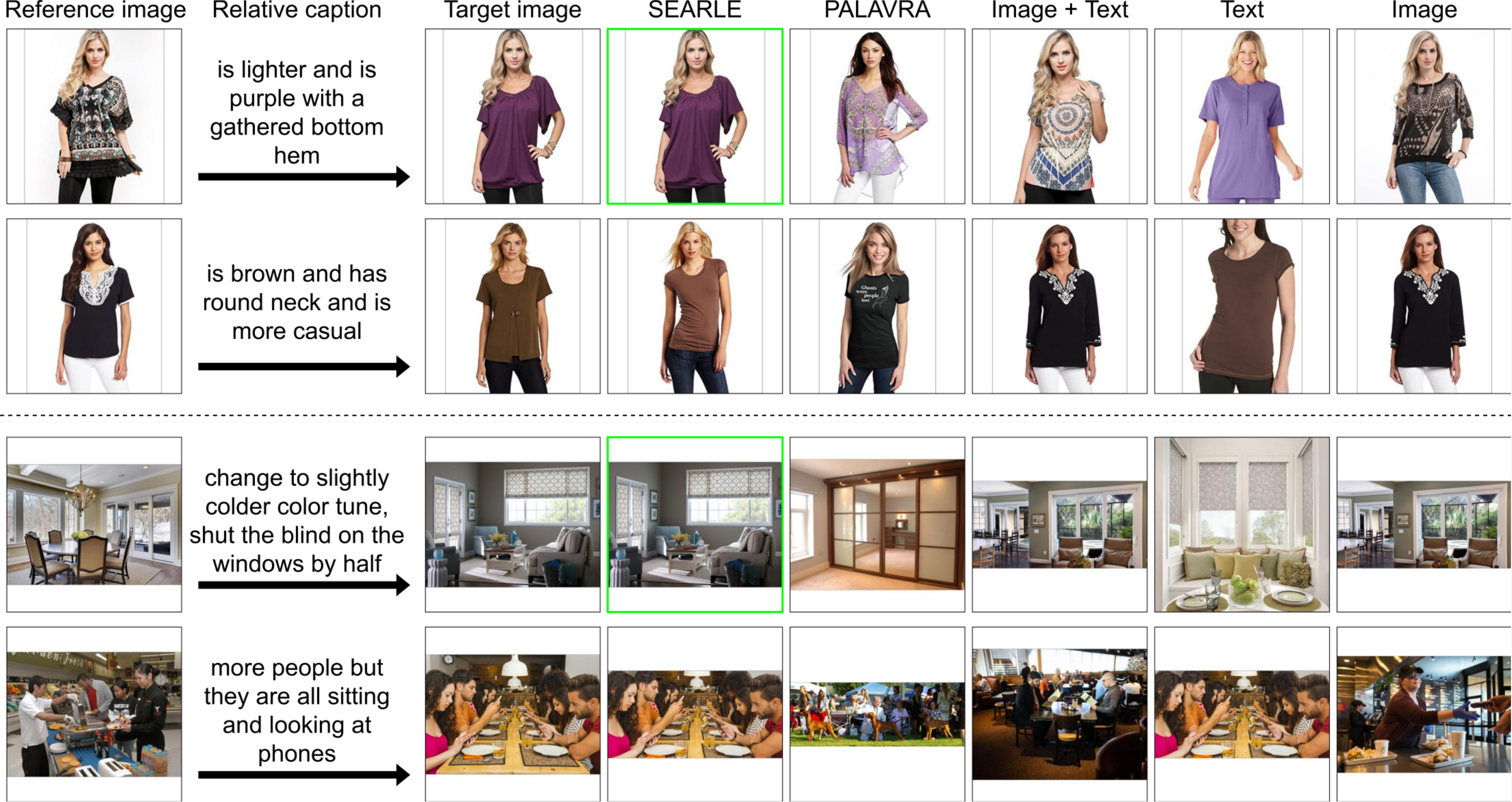}
    \caption{Qualitative results for the FashionIQ (top) and CIRR (bottom) datasets. For a clearer visualization, we do not consider the reference image in the retrieval results. We highlight with a green border when the retrieved image is the labeled ground truth. The second and fourth rows show examples in which \method retrieves a false negative. \vspace{15pt}}
    \label{fig:qualitative_cirr_fiq}
\end{figure*}

\Cref{fig:qualitative_cirr_fiq} shows the qualitative results for FashionIQ and CIRR. We observe that \method manages to integrate the visual features of the reference image and the text features of the relative caption to retrieve the correct image. On the contrary, the baselines either focus too much on the reference image or the relative caption. The second and fourth rows of the figure highlight the problem of false negatives in existing CIR datasets. Indeed, \method retrieves images that are valid matches for the query but are not labeled as such.

\begin{figure*}
    \centering
    \includegraphics[width=0.97\textwidth]{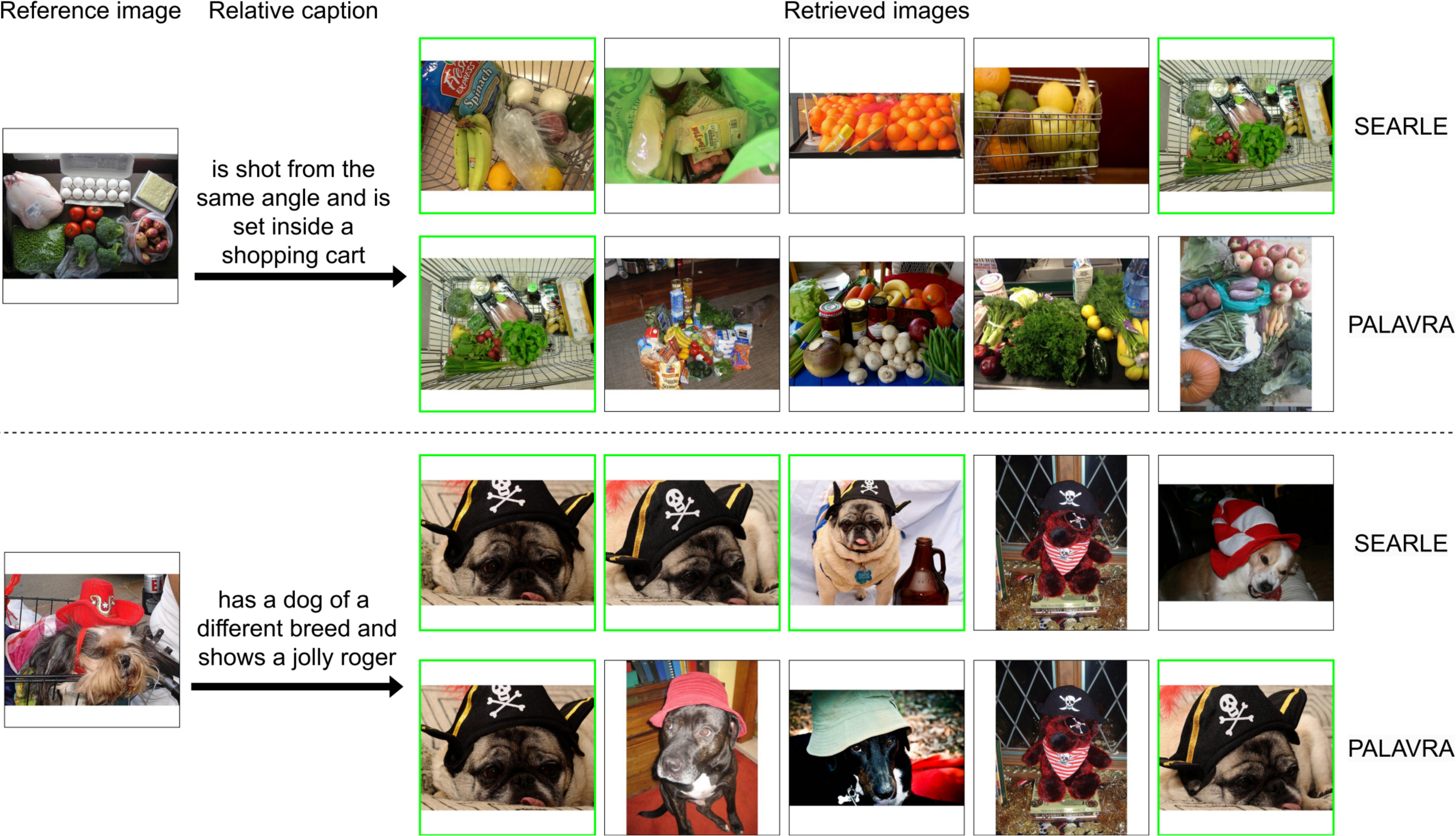}
    \caption{Qualitative results for the CIRCO dataset. We compare the top-5 retrieved images of \method and the best-performing baseline. We highlight ground truths with a green border.}
    \label{fig:qualitative_circo}
\end{figure*}

In \cref{fig:qualitative_circo} we compare the top-5 images retrieved by \method and PALAVRA for two queries belonging to CIRCO. \method manages to retrieve more relevant images compared to PALAVRA. We stress that without the second phase of the annotation process (see \cref{sec:multiple_gt_annotation}) the additional ground truths would have been false negatives.

\end{document}